\documentclass{article} % For LaTeX2e
\usepackage[dvipsnames,svgnames]{xcolor}
\usepackage{iclr2025_conference,times}

\usepackage[authoryear,round,sort,compress]{natbib}
\usepackage{booktabs}
\usepackage{multirow}	
\definecolor{MyRef}{HTML}{74787c}     %
\usepackage{tikz}
\usepackage{lineno} 
\usepackage[toc,page]{appendix}
\usepackage[colorlinks,linkcolor=MyRef,anchorcolor=green,citecolor=blue]{hyperref}
\definecolor{darkblue}{HTML}{1A254B}
\definecolor{lightblue}{HTML}{A7BED3}
\definecolor{blue}{HTML}{114083}
\definecolor{blue2}{HTML}{0000ff}

\usepackage{amsfonts,amssymb}
\usepackage{amsthm}
\usepackage{threeparttable}
\usepackage{wrapfig}
\usepackage{subfigure}
\usepackage{algpseudocodex,algorithm}
\usepackage{amsmath}
\usepackage{graphicx}
\usepackage{cite}
\newcommand{\revise}[1]{\textcolor{black}{#1}}
\newcommand{\revisenew}[1]{\textcolor{black}{#1}}

\usepackage{enumerate}
\usepackage[sort&compress,capitalize,nameinlink]{cleveref}

\newcommand{\eqqref}[1]{Eq. (\ref{#1})}

\title{Learning stochastic dynamics from snapshots through regularized unbalanced optimal transport}

\author{Zhenyi ~Zhang$^1$, ~  Tiejun Li$^{1, 2, \ast}$, and ~  Peijie Zhou $^{2,3,4,5,}$\thanks{Corresponding author.} \\
$^1$LMAM and School of Mathematical Sciences, Peking University.\\
$^2$Center for Machine Learning Research, Peking University. $^3$NELBDA, Peking University.  \\ $^4$Center for Quantitative Biology, Peking University. $^5$AI for Science Institute, Beijing.
\\
\texttt{zhenyizhang@stu.pku.edu.cn}, \texttt{\{tieli, pjzhou\}@pku.edu.cn}
}
\newtheorem{theorem}{\noindent Theorem}[section]
\newtheorem{definition}{\noindent Definition}[section]
\newtheorem{remark}{\noindent Remark}[section]

\newcommand{\Xb}{{\boldsymbol X}}
\newcommand{\Yb}{{\boldsymbol Y}}
\newcommand{\xb}{{\boldsymbol x}}

\newcommand{\Rd}{\mathbb{R}^d}
\def\rmdx{\mathrm{d} \boldsymbol{x}}

\def\pxt{{p}(\boldsymbol{x},t)}
\def\bxt{\boldsymbol{b}(\boldsymbol{x},t)}
\def\vxt{\boldsymbol{v}(\boldsymbol{x},t)}
\def\gxt{{g}(\boldsymbol{x},t)}
\def\tgxt{\widetilde{g}(\boldsymbol{x},t)}
\def\dxt{\boldsymbol{\sigma}(\boldsymbol{x},t)}

\def\dxT{\boldsymbol{\sigma}^{T}(\boldsymbol{x},t)}
\def\axt{\boldsymbol{a}(\boldsymbol{x},t)}
\def\ainvxt{\boldsymbol{a}^{-1}(\boldsymbol{x},t)}
\def\aXt{\boldsymbol{a}({\boldsymbol X}_t,t)}

\def\Xt{{\boldsymbol X}_t}
\def\Wt{{\boldsymbol W}_t}
\def\Yt{{\boldsymbol Y}_t}
\def\Ws{{\boldsymbol W}_s}
\def\Ys{{\boldsymbol Y}_s}
\def\rmd{\mathrm{d}}

\iclrfinalcopy % Uncomment for camera-ready version, but NOT for submission.
\begin{document}
\maketitle

\begin{abstract}
Reconstructing dynamics using samples from sparsely time-resolved snapshots is an important problem in both natural sciences and machine learning. Here, we introduce a new deep learning approach for solving regularized unbalanced optimal transport (RUOT) and inferring continuous unbalanced stochastic dynamics from observed snapshots. Based on the RUOT form, our method models these dynamics without requiring prior knowledge of growth and death processes or additional information, allowing them to be learned directly from data.  Theoretically, we explore the connections between the RUOT and Schrödinger bridge problem and discuss the key challenges and potential solutions. The effectiveness of our method is demonstrated with a synthetic gene regulatory network, high-dimensional Gaussian Mixture Model, and single-cell RNA-seq data from blood development. Compared with other methods, our approach accurately identifies growth and transition patterns, eliminates false transitions, and constructs the Waddington developmental landscape. Our code is available at: \href{https://github.com/zhenyiizhang/DeepRUOT}{https://github.com/zhenyiizhang/DeepRUOT}.
\end{abstract}

\section{Introduction}

In machine learning and natural sciences, a key challenge is coupling high-dimensional distributions from observed samples, exemplified by Variational Autoencoders (VAEs) \citep{kingma2022autoencodingvariationalbayes} which map complex data to simpler latent spaces. It is also important in multi-modal analysis for integrating diverse data types \citep{lahat2015multimodal}, particularly in biology through aligning multi-omics data into unified cellular state representations \citep{demetci2022scot,cao2022unified,gao2024graspot,cang2024synchronized}. Recently, there has been growing interest in understanding the dynamics of how distributions are coupled over time, such as diffusion models \citep{ho2020denoising,sohl2015deep} and stochastic differential equations (SDEs) \citep{song2020score}. The task is useful for interpolating paths between arbitrary distributions and learning the underlying dynamics \citep{sflowmatch,action_matching,de2021diffusion,wang2021deep}. 

Due to the destructive nature of technology, the analysis of time series data in single-cell RNA sequencing (scRNA-seq) provides an important application scenario for the high dimensional probability distribution coupling and dynamical inference problem \citep{waddingot,Tigon,peng2024stvcr,trajectory,jiang2022dynamic,jiang2024physics,bunne2024optimal,bunne2023learning,tong2023unblanced,Stephenzhang2021optimal}. Trajectory inference in scRNA-seq data have been extensively studied \citep{saelens2019comparison}, and optimal transport (OT)-based methods has emerged as a central tool for datasets with temporal resolution  \citep{bunne2024optimal,waddingot,moscot,genot}. Often, there is a need to learn the continuous dynamics of cells over time and fit the mechanistic model that transforms the initial cell distributions into the distributions at later temporal points. This could be solved through the dynamical formulation of OT  \citep{benamou2000computational}, also known as the B-B form.  However, the formulation has not fully taken the stochastic dynamical effects into account, especially the intrinsic noise in gene expression and cell differentiation \citep{zhou2021stochasticity,zhou2021dissecting}, which is prevalent in biological processes on single-cell level \citep{elowitz2002stochastic}.

By incorporating stochastic dynamics, the Schrödinger bridge (SB) problem aims to identify the most likely stochastic transition path between two arbitrary distributions relative to a reference stochastic process \citep{sb}, and  has been applied in a wide range of contexts, including 
 scRNA-seq analysis and generative modeling   \citep{bunne_unsb,Liudeep_genera_sb}.  Meanwhile, recent \textbf{regularized unbalanced optimal transport (RUOT)} offers a promising approach for modeling both stochastic unbalanced continuous dynamics \citep{chen2022most,branRUOT,highorderRUOT,janati2020entropic}, which can be viewed as an unbalanced relaxation of the dynamic formulation of Schrödinger bridge problem. 
 \revise{However, computational methods for learning RUOT or such high-dimensional unbalanced stochastic dynamics from snapshots are relatively lacking, especially when there is no prior knowledge of unbalanced effect.}

Here we develop a new deep learning method (DeepRUOT)
%\footnote{Code is avaliable at \href{https://anonymous.4open.science/r/DeepRUOT-859F/}{https://anonymous.4open.science/r/DeepRUOT-859F/}}
 for learning general RUOT and inferring continuous unbalanced stochastic dynamics from samples based on the derived Fisher regularization form without requiring prior knowledge. We demonstrate the effectiveness of DeepRUOT on both synthetic and real-world datasets. Compared to the common SB method, our approach accurately identifies growth and transition patterns, eliminates false transitions, and constructs the Waddington developmental landscape of scRNA-seq data. Overall, our main contributions can be summarized as follows:
\begin{itemize}
\item We reformulate RUOT with a Fisher regularization form and explore the connections between RUOT and unbalanced SB. The formulation transforms the SDE  into the ordinary differential equation (ODE), which is computationally more tractable. 

\item We propose DeepRUOT, the neural network algorithm for learning high dimensional unbalanced stochastic dynamics from snapshots. \revise{Through the neural network modeling for growth and death, our framework does not require prior knowledge of these processes.} 
    
\item We validate the effectiveness of DeepRUOT on both synthetic data and real scRNA-seq datasets, showing its promising performance compared with existing approaches.
\end{itemize}

\section{Related Works}

\paragraph{Deep Learning Solver for Dynamical OT} To tackle the dynamical OT (i.e. B-B form) in high dimensions, many methods \citep{trajectorynet,mioflow,scnode,Liwu2020machine,zhouhaolearning,shenjie2024new,DOngbin2023scalable,pooladian2024neural,cfm_lipman,cfm_tong,albergo2023building,liu2023flow,jiao2024convergenceanalysisflowmatching,jiao2024convergencecontinuousnormalizingflows,zhouhao2022neural,zhouhao2024parameterized_H,zhouhao2024parameterized_W,LiwuZHou2020wasserstein,Lujianfeng2024convergence} have been developed based on continuous normalizing flow and neural ODE formulation either in original or latent space. To account for sink and source terms in unnormalized distributions,  \citep{Tigon,peng2024stvcr,tong2023unblanced} formulated the neural-network based solver in the unbalanced dynamical OT setup. However, the appropriate formulation along with an effective deep learning solver to \textbf{simultaneously account for unbalanced term and stochastic effects} in dynamical OT remains largely lacking. 

\paragraph{Computational Methods for Schrödinger Bridge Problem}
Many methods have recently been developed to solve the static SB problem \citep{trajectory,bunne_unsb,Geofftrajectory,Geofftrajectory2,shi2024diffusion,de2021diffusion,pooladian2024plug,Liudeep_genera_sb,gu2024partially}. To tackle the dynamical Schrödinger Bridge, methods based on neural SDE solver, neural ODE solver with fisher information regularization or flow matching \citep{dyn_sb_koshizuka2023neural,action_matching,tong_action,Duan_action,bunne_dynam_SB,chen2022likelihood,albergo2023stochastic,wang2021deep,Jiao2024LatentSB,zhou2024denoising,imagesb,LIWUCHEN_score,sflowmatch,maddu2024inferring} have been proposed. However, these methods either fail to account for \textbf{unnormalized distributions} resulting from cell growth and death, or require \textbf{prior knowledge} of the processes (growth/death rate) \citep{waddingot,trajectory,bunne_unsb,Geofftrajectory2} or additional information (e.g.cell lineage) \citep{Geofftrajectory}.

\paragraph{Study of RUOT}
The RUOT has recently formulated \citep{chen2022most}, also known as unbalanced Schrödinger Bridge. The existing studies are mostly focused on \textbf{the theoretical side}. For instance, \citep{janati2020entropic} derived  a closed-form formula for entropic OT between unbalanced Gaussian measures. \citep{branRUOT} investigates  relations between RUOT with branching Schrödinger bridge. \citep{highorderRUOT} investigates different formulations of the problem.

\section{Preliminaries and Backgrounds}
\revise{In this section, we provide an overview of stochastic effects and unbalanced forms within the dynamical framework. Specifically, considering only stochastic effects leads to the Schrödinger bridge problem (\cref{sec:SBP}), whereas addressing solely the unbalanced aspect results in unbalanced dynamic optimal transport (\cref{sec:undynot}). By integrating these two perspectives, we motivate the formulation of the Regularized Unbalanced Optimal Transport (RUOT) framework.}
\subsection{\revise{Stochastic Effect: Schrödinger Bridge Problem}}\label{sec:SBP}
The Schrödinger bridge problem aims to identify the most likely evolution between a given initial distribution $\nu_0$ and a terminal distribution $\nu_1$ (assumed to have density in this paper), relative to a given reference stochastic process. Formally, this problem can be formulated as the minimization of the Kullback-Leibler (KL) divergence in the optimal control perspective \citep{dai1991stochastic} as below:
\begin{equation}
\label{eq:sb}
    \min_{\mu^\Xb_0=\nu_0,~  \mu^\Xb_1=\nu_1} \mathcal{D}_{\mathrm{KL}}\left(\mu^\Xb_{[0,1]} \| \mu^\Yb_{[0,1]}\right),
\end{equation}
where $\mu^\Xb_{[0,1]}$ denotes the probability measure induced by stochastic process $\Xt \ \left(0\leq t\leq 1\right)$ defined on the space of all continuous paths $C([0,1],\mathbb{R}^d)$, with the distribution of $\Xt$ at given time $t$ characterized by the measure $\mu^\Xb_{t}$ with density function $p(\boldsymbol{x},t)$. In this work, we consider $\Xt$ the stochastic process characterized by the following stochastic differential equation (SDE): 
\begin{equation}\label{eq:sde}
    \mathrm{d} \Xt=\boldsymbol{b}\left(\Xt, t\right) \mathrm{d} t+\boldsymbol{\sigma}\left(\Xt, t\right)\mathrm{d}\Wt,
\end{equation}
and the reference measure
$\mu^\Yb_{[0,1]}$ is chosen as the probability measure induced by the process \(\rmd \boldsymbol{Y}_t = \boldsymbol{\sigma}(\boldsymbol{Y}_t, t) \rmd \Wt\), where $\Wt \in \mathbb{R}^d$ is the standard multidimensional Brownian motion defined on a probability space $(\Omega,\mathcal{F},\mathbb{P})$ with $\mathbb{P}$ the Wiener measure for the coordinate process $\Wt(\omega)=\omega(t)$, and $\boldsymbol{\sigma}(\boldsymbol{x}, t) \in \mathbb{R}^{d\times d}$ is the diffusion matrix which is typically assumed as bounded, coercive and invertible. Under this formulation, the solution to \eqqref{eq:sb} is also referred to as the \emph{diffusion Schrödinger bridge}. The diffusion Schrödinger bridge also possesses a dynamic formulation, which can be formally stated as the following theorem.
\begin{theorem}
\label{thm:thm_fisher_simple}
    Consider the diffusion Schrödinger bridge \revise{problem}
  (\ref{eq:sb})
    where $\mu^\Yb_{[0,1]}$ the reference measure induced by \(\rmd \boldsymbol{Y}_t = \boldsymbol{\sigma}(\boldsymbol{Y}_t, t) \rmd \Wt\). Then (\ref{eq:sb}) is equivalent to 
    \begin{equation}
        \label{eq:dsb2}
\inf _{\left({p}, \boldsymbol{b}\right)} \int_0^1\int_{\mathbb{R}^d}\left[\frac{1}{2}\boldsymbol{b^T}(\boldsymbol{x}, t)\ainvxt\boldsymbol{b}(\boldsymbol{x}, t)\right]\pxt\rmd \boldsymbol{x}\mathrm{d} t ,
    \end{equation}
where the infimum is taken over all function pairs $\left({p}, \boldsymbol{b}\right)$ such that ${p}(\cdot , 0)=$ ${\nu_0}$, ${p}(\cdot , 1)={\nu_1}$, ${p}(\boldsymbol{x},t)$ absolutely continuous, and
    \begin{equation}
        \label{eq:dsb3}
\partial_t \pxt=-\nabla_{\boldsymbol{x}} \cdot\left(\pxt \bxt\right)+\frac{1}{2} \nabla_{\boldsymbol{x}}^2:\left(\axt \pxt\right),
    \end{equation}
    where $\nabla_{\boldsymbol{x}}^2:\left(\axt \pxt\right)=\sum_{ij}\partial_{ij}(a_{ij}p)$ and $\axt=\dxt \dxT$,
    coupled with vanishing boundary condition: $\displaystyle \lim_{|\boldsymbol{x}| \rightarrow \infty}p(\boldsymbol{x},t)=0$.
\end{theorem}

This theorem or its variants have been stated and proven in various forms, such as in \citep{sb_Gentil,sb_chen,dai1991stochastic}. Here, we provide a simple direct proof for illustration in  \cref{appen:proofdynaSB}.
\revise{We note that in this dynamic formulation, it accounts for the stochastic aspects but does not incorporate unbalanced effects. As we will discuss below, RUOT can be motivated by incorporating unbalanced effects in the dynamic formulation of SB.}
\subsection{\revise{Unbalanced Effect: Unbalanced Dynamic Optimal Transport}}\label{sec:undynot}
The optimal transport problem has been extensively studied in various fields.  Given two probability distributions \(\boldsymbol{\alpha} \in \mathbb{R}_+^n\) and \(\boldsymbol{\beta} \in \mathbb{R}_+^m\), its primary goal is to  find the optimal coupling \(\boldsymbol{\pi} \in \mathbb{R}_+^{n \times m}\) to transport a given distribution of mass (or resources) from one location to another while minimizing the cost associated with the transportation. The static optimal transport problem can be mathematically formulated as
$\min_{\boldsymbol{\pi} \in \boldsymbol{\Pi}(\boldsymbol{\alpha}, \boldsymbol{\beta})}\langle\boldsymbol{\pi}, \boldsymbol{c}\rangle$, where 
$\boldsymbol{\Pi}(\boldsymbol{\alpha}, \boldsymbol{\beta})=\left\{\boldsymbol{\pi} \in \mathbb{R}_{+}^{n \times m}: \boldsymbol{\pi} \mathbf{1}_m=\boldsymbol{\alpha}, \boldsymbol{\pi}^T \mathbf{1}_n=\boldsymbol{\beta}, \boldsymbol{\pi} \geq 0\right\}.$
The cost matrix \(\boldsymbol{c} \in \mathbb{R}^{n \times m}\) defines the transportation cost between each pair of points, where \(c_{ij}:= c(\boldsymbol{x}_i, \boldsymbol{y}_j)\) represents the cost of moving a unit mass from point \(\boldsymbol{x}_i\) to point $\boldsymbol{y}_j$.
We refer to \citep{ot_theory} for more details. Next we briefly state some well-known results on the dynamical formulation of the OT.

\paragraph{Dynamical Optimal Transport} The formulation is also known as the Benamou-Brenier formulation \citep{benamou2000computational}, which can be stated as follows: 
$$
\begin{gathered}
\mathcal{W}\left(\nu_0, \nu_1\right)=\inf _{\left(\pxt, \bxt\right)} \int_0^1 \int_{\mathbb{R}^d} \frac{1}{2}\|\bxt\|_2^2 \pxt\rmd \boldsymbol{x} \rmd t, \\
\mbox{s.t. }\partial_t p+\nabla \cdot\left(\bxt p\right)=0,\ p|_{t=0}=\nu_0, p|_{t=1}=\nu_1 .
\end{gathered}
$$
Compared to SB, in OT the distributions are connected by deterministic transport equation instead of diffusion. It can be shown that this dynamical formulation corresponds to a static Kantorovich's optimal transport problem with  cost function $c(\boldsymbol{x},\boldsymbol{y})=\|\boldsymbol{x}-\boldsymbol{y}\|_2^2$.
\paragraph{Regularized Optimal Transport}
The regularized optimal transport is defined by a dynamical form of the general Schrödinger bridge by taking diffusion rate as constant and scaled by $\sigma^2$ in \eqref{eq:dsb2}:
$$
\begin{gathered}
\mathcal{W}\left(\nu_0, \nu_1\right)= \inf _{\left(\pxt, \bxt\right)} \int_0^1 \int_{\mathbb{R}^d} \frac{1}{2}\|\bxt\|_2^2  \pxt\rmd \boldsymbol{x} \mathrm{d} t, \\
\mbox{s.t. } \partial_t p+\nabla \cdot\left(\bxt p\right)-\frac{\sigma^2}{2} \Delta p=0,\ p|_{t=0}=\nu_0, p|_{t=1}=\nu_1,
\end{gathered}
$$
It can be demonstrated that as  $\sigma^2$  approaches zero, the solution to this problem converges to that of the Benamou-Brenier problem \citep{rot}. \revise{And it is equivalent to the SB problem \citep{sb_Gentil,branRUOT,sb}}.
\paragraph{Unbalanced Dynamic Optimal Transport}
To account for unnormalized marginal distributions and effects such as growth and death, an unbalanced optimal transport problem with Wasserstein–Fisher–Rao (WFR) metric has been proposed \citep{uot1,uot2} or its extensions \citep{liunnormalized}. Here we adopt the WFR unbalanced optimal transport:
$$
\begin{gathered}
\mathcal{W}\left(\nu_0, \nu_1\right)=\inf _{\left(\pxt, \bxt,\gxt\right)} \int_0^1 \int_{\mathbb{R}^d} \left(\frac{1}{2}\|\bxt\|_2^2 +\alpha |\gxt|_2^2\right)  \pxt \rmd \boldsymbol{x} d t, \\
\mbox{s.t. }\partial_t p+\nabla \cdot\left(\bxt p\right)=g(\boldsymbol{x},t) p,\ p|_{t=0}=\nu_0, p|_{t=1}=\nu_1 .
\end{gathered}
$$
Here $g(\boldsymbol{x},t)$ is a scalar function that denotes the growth or death rate of particles at the state $\boldsymbol{x}$ and time $t$, and is also optimized in the total energy term. $\alpha$ is the hyperparameter of weight. We should also note that in this case, $\nu_0$ and $\nu_1$ are not necessarily the normalized probability densities, but are generally densities of masses.
\section{Regularized Unbalanced Optimal Transport}\label{sec:ruot}

To simplify the notation and illustrate a commonly used setting, we take a special case considering  $\axt=\sigma^2(t)\boldsymbol{I}$, and the general case is left to \cref{appen:RUOTgeneral} for further discussions. Inspired by unbalanced dynamic optimal transport, the dynamical formulation \cref{thm:thm_fisher_simple} suggests a natural approach to relaxing the mass conservation constraint by introducing a growth/death term $gp$ in \eqqref{eq:dsb3} as 
$        
\partial_t p=-\nabla_{\boldsymbol{x}} \cdot\left(p \boldsymbol{b}\right)+\frac{1}{2}\nabla_{\boldsymbol{x}}^2: \left(\sigma^2(t)\boldsymbol{I}p\right)+gp.
$
   Meanwhile, we also define a loss functional in \eqqref{Eq:3.3} which incorporates the growth penalization and Wasserstein metric considered in \eqqref{eq:dsb2} (with a rescaling by $\sigma^2(t)$).  We refer to this formulation as the 
    \emph{regularized unbalanced optimal transport}.

    \begin{definition}[Regularized unbalanced optimal transport]
    \label{def:unddsb}
     Consider
       \begin{equation}\label{Eq:3.3}
\inf _{\left(p, \boldsymbol{b}, g\right)} \int_0^1 \int_{\Rd}\frac{1}{2}\left\|\bxt\right\|_2^2\pxt\rmd \boldsymbol{x}\mathrm{d} t+ \int_0^1 \int_{\Rd} \alpha \Psi\left(g(\boldsymbol{x},t)\right)\pxt    \rmdx\mathrm{d} t,
    \end{equation}
    where \revise{$\Psi: \mathbb{R}\rightarrow [0, +\infty]$ corresponds to the growth penalty function,} the infimum is taken over all pairs $\left(p, \boldsymbol{b}, g\right)$ such that $p(\cdot , 0)=$ $\nu_0, p(\cdot , 1)=\nu_1, p(\boldsymbol{x},t)$ absolutely continuous, and
    \begin{equation}
        \label{eq:undsb3}
\partial_t p=-\nabla_{\boldsymbol{x}} \cdot\left(p \boldsymbol{b}\right)+\frac{1}{2}\nabla_{\boldsymbol{x}}^2: \left(\sigma^2(t)\boldsymbol{I}p\right)+gp 
    \end{equation}
    with vanishing boundary condition: $\displaystyle \lim_{|\boldsymbol{x}| \rightarrow \infty}p(\boldsymbol{x},t)=0$.
    \end{definition}
    
Note that in the definition if $\Psi(g)=+\infty$ unless $g=0$ and $\Psi(0)=0$, then it implies $g(\boldsymbol{x},t)=0$ and the RUOT degenerates to the \emph{regularized optimal transport} problem. If $\sigma(t) \rightarrow 0$ and $\Psi\left(g(\boldsymbol{x},t)\right)=|g(\boldsymbol{x},t)|^2$, this degenerates to the \emph{unbalanced dynamic optimal transport} with WFR metrics. Meanwhile when $\sigma(t)$ is  constant, it coincides with the definition of RUOT provided in \citep{branRUOT}. Then, we can reformulate \cref{def:unddsb} with the following Fisher information regularization. 
\begin{theorem}
\label{thm:unddsb_fisher}
The  regularized unbalanced optimal transport problem \eqref{Eq:3.3} is equivalent to 
\begin{equation}
\label{eq:unddsb_fisher}
\inf _{\left(p, \boldsymbol{v}, g\right)} \int_0^1\int_{\Rd}\left[ \frac{1}{2}\left\|\boldsymbol{v}\right\|_2^2+\frac{\sigma^4(t)}{8}\left\|\nabla_{\boldsymbol{x}} \log p\right\|_2^2-\frac{\sigma^2(t)}{2}\left(1+\log p\right)g-\frac{1}{2}\frac{\rmd \sigma^2(t)}{\rmd t}\log p+ \alpha \Psi\left(g\right) \right]p\rmdx\mathrm{d} t,
\end{equation}
where the infimum is taken over all triplets $\left(p, \boldsymbol{v}, g\right)$ such that $p(\cdot , 0)=$ $\nu_0, p(\cdot , 1)=\nu_1, p(\boldsymbol{x},t)$ absolutely continuous, and
\begin{equation}
\label{eq:unddsb_fisher_contineq}
    \partial_t p=-\nabla_{\boldsymbol{x}} \cdot\left(p \vxt\right) + g(\boldsymbol{x},t) p 
\end{equation}
with vanishing boundary condition: $\displaystyle \lim_{|\boldsymbol{x}| \rightarrow \infty}p(\boldsymbol{x},t)=0$.
 \end{theorem}

  \revise{Here $\vxt$ represents a new vector field.} The proof is left to \cref{appen:fisherRUOT}. \revisenew{In \citep{branRUOT}, they proposed another Fisher regularization form of RUOT, expressed in the formula \cref{eqn:laventfisher}. \revisenew{In their formulation, computing the cross term $\langle \nabla_{\boldsymbol{x}} \log p, \sigma^2(t) \boldsymbol{v} \rangle$ required calculating the derivative of the function $\log p$ with respect to $\boldsymbol{x}$ and performing vector multiplications with $\boldsymbol{v}$.} The formulation here is equivalent to theirs but more computationally tractable, as we avoid differentiation and vector multiplication in our cross-term.}
  \begin{remark}
     When $g=0$, $\Psi(0)=0$ and $\sigma(t)$ is constant, then \eqqref{eq:unddsb_fisher} is the same as the dynamic entropy-regularized optimal transport form as discussed in \citep{bunne_dynam_SB,li2020fisher,sb_Gentil,pooladian2024plug,tong_action,LiWU2021schb}.
 \end{remark}
\begin{remark}
 The  term  
 $
 \mathcal{I}(p)=\int_{\Rd} \|\nabla_{\boldsymbol{x}} \log \pxt \|_2^2 p(\boldsymbol{x},t) \rmdx
 $
 in \eqqref{eq:unddsb_fisher} is referred to as the \emph{Fisher information}. Notably, when considering growth/death factors, \eqqref{eq:unddsb_fisher} includes not only the Fisher-Rao metric but also an additional cross-term
 $
 \int_{\Rd}-\frac{1}{2}\sigma^2(t)\left(1+\log \pxt\right)g(\boldsymbol{x},t) p(\boldsymbol{x},t) \rmdx.
 $
\end{remark}
\begin{remark}
  From the Fokker-Plank equation and the proof of \cref{thm:unddsb_fisher},  the original SDE
  $\mathrm{d} \Xt=\left(\boldsymbol{b}\left(\Xt, t\right)\right) \mathrm{d} t+{\sigma}\left( t\right)\mathrm{d}\Wt$
  can be transformed 
  into the probability flow ODE
  $$
  \mathrm{d} \Xt=\underbrace{\left(\boldsymbol{b}\left(\Xt, t\right)-\frac{1}{2} \sigma^2(t)\nabla_{\boldsymbol{x}} \log p(\Xt,t)\right)}_{\boldsymbol{v}\left(\Xt, t\right)} \rmd t.
  $$
  Conversely, if the probability flow ODE’s drift $\vxt$, the diffusion rate $\sigma(t)$ and the \emph{score function} $\nabla_{\boldsymbol{x}} \log p(\boldsymbol{x},t)$ are known, then the
the drift term $\boldsymbol{b}(\boldsymbol{x},t)$ of the SDE can be determined by
$
\boldsymbol{b}(\boldsymbol{x},t)=\vxt+\frac{1}{2}\sigma^2(t) \nabla_{\boldsymbol{x}} \log p(\boldsymbol{x},t).
$
\revise{Thus, to specify an SDE is equivalent to specifying the probability flow ODE and the corresponding score function $\nabla_{\boldsymbol{x}} \log \pxt$ \citep{sflowmatch}}.    
\end{remark}

  In \citep{highorderRUOT}, the authors defined a RUOT problem with nonlinear Fokker-Planck equation constraints. In \cref{appen:RUOTnonlinear}, we show that the proposed form is indeed consistent with the RUOT defined here when $\sigma(t)$ is constant and $\Psi\left(\gxt\right)$ takes the quadratic form (i.e., $\Psi\left(\gxt\right)=|\gxt|_2^2$).   We then  explore the connections between RUOT and SB problem in \cref{appen:connectionSB}.

\section{Learning  RUOT  through Neural Networks}\label{sec:deepruot}

\begin{wrapfigure}{R}{0.45\textwidth}
     \begin{minipage}{0.45\textwidth}
     \vspace{-3em}
        \begin{figure}[H]
            \centering
            \includegraphics[width=1\textwidth]{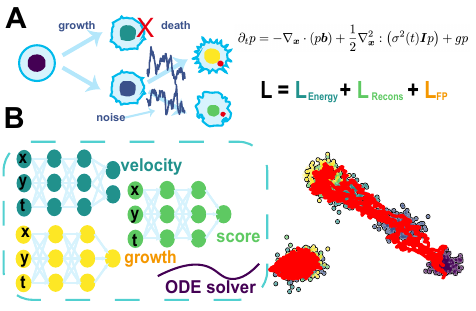}
            \caption{\revise{Overview of DeepRUOT.}}
        \label{fig:DeepRUOT}
        \end{figure}
        \vspace{-1em}
     \end{minipage}
\end{wrapfigure}

Given unnormalized distributions at $T$ discrete time points, $\boldsymbol{X}_i \sim \mu_i$ for fixed timepoints $i \in \{0, \ldots, T-1\}$, we aim to learn continuous stochastic dynamics satisfying the RUOT from data. Similarly, to simplify the exposition and illustrate a commonly used case, we consider  \cref{def:unddsb} with the stochastic dynamics 
$
\mathrm{d} \Xt=\boldsymbol{b}\left(\Xt, t\right) \mathrm{d} t+{\sigma}\left( t\right)\mathrm{d}\Wt.
$
As previously discussed, we approach this by transforming the problem into learning the drift $\vxt$ of the probability ODE and its score function $\frac{1}{2}\sigma^2(t) \nabla_{\boldsymbol{x}}\log p(\boldsymbol{x},t)$ \revise{in \cref{thm:unddsb_fisher}}.  We parameterize $\vxt$, $\gxt$, and $\frac{1}{2}\sigma^2(t)\log \pxt$ using neural networks $\boldsymbol{v}_\theta, ~ g_\theta$ and $s_\theta$ respectively (\cref{fig:DeepRUOT}). \revise{To solve \cref{thm:unddsb_fisher}, the overall loss is composed of energy loss, reconstruction loss, and the Fokker-Planck constraint:
\begin{equation}\label{eq:total_loss}
\mathcal{L}= \mathcal{L}_{\text{Energy}}+\lambda_r\mathcal{L}_{\text{Recons}}+\lambda_f\mathcal{L}_{\text{FP}}.
\end{equation}
The $\mathcal{L}_{\text{Energy}}$ loss promotes the least action of kinetic energy \eqqref{eq:unddsb_fisher}. The reconstruction loss $\mathcal{L}_{\text{Recons}}$ promotes the  dynamics to match
 data distribution at later time point (i.e. $p(\cdot, 1)=\nu_1$), and the $\mathcal{L}_{\text{FP}}$ promotes the three parameterized neural network to satisfy Fokker-Planck constraints \eqqref{eq:unddsb_fisher_contineq}.
}

\subsection{Energy Loss} To compute the integral in \eqqref{eq:unddsb_fisher}, the direct calculation is infeasible due to the high dimensionality. Thus, we need to transform it into an equivalent form that can be evaluated using Monte Carlo methods. 
\revise{Adopting  the approach in \citep{Tigon} (see \cref{appendix:energy}),  \eqqref{eq:unddsb_fisher} is equivalent to the following form}
\revise{
\begin{equation}
    \label{eq:unddsb_cost}
    \begin{aligned}
\mathcal{L}_{\text{Energy}}=&\mathbb{E}_{\boldsymbol{x_0} \sim p_0}\int_0^T\left[ \frac{1}{2}\left\|\boldsymbol{v}_\theta\right\|_2^2+\frac{1}{2}\left\|\nabla_{\boldsymbol{x}} s_\theta\right\|_2^2-\left(\frac{\sigma^2(t)}{2}+s_\theta\right)g_\theta- \frac{(\sigma^2(t))'}{\sigma^2(t)}s_\theta+ \alpha \Psi\left(g_\theta\right) \right] w_\theta (t)\mathrm{d} t,
    \end{aligned}
\end{equation}
where $w_\theta(t)=e^{\int_0^t g_\theta(\boldsymbol{x}(t), s) \rmd s}$ and $\boldsymbol{x}(t)$ satisfy $\rmd \boldsymbol{x}/\rmd t =\vxt \rmd t$. We compute through Monte Carlo sampling and a Neural ODE solver.}

\subsection{Reconstruction Loss}

The reconstruction loss aims to match the final distribution in \cref{thm:unddsb_fisher} \revise{(i.e., $p(\cdot , 1)=\nu_1$). Many works use the balanced optimal transport to evaluate the distance between two distributions. However due to the unnormalized effect, here we aim to use  the unbalanced optimal transport
instead. To realize this we need to 
tackle  two parts: 
\begin{equation}\label{eq:unddsb_recon}
\mathcal{L}_{\text{Recons}}=\lambda_m\mathcal{L}_{\text{Mass}}+\lambda_d\mathcal{L}_{\text{OT}}
\end{equation}
where $\lambda_m$ and $\lambda_d$ are hyperparameters. The mass matching loss $\mathcal{L}_{\text{Mass}}$ promotes to align the number of cells. We then normalize the distributions according to the matched masses. The $\mathcal{L}_{\text{OT}}$ uses these weights to perform optimal transport matching. }

For $\mathcal{L}_{\text{Mass}}$, we propose a local mass matching strategy.
We denote $A_i$ ($i=0,1, \cdots T-1$) as the dataset observed at different time points, and define $\revisenew{\phi_\theta^{\boldsymbol{v}}}: \mathbb{R}^d \times \mathcal{T} \rightarrow \mathbb{R}^{d|\mathcal{T}|}$ as the \revisenew{trajectory mapping function of the Neural ODE $\rmd \boldsymbol{x}/\rmd t=\boldsymbol{v}_\theta$, which inputs a given starting point as the initial condition and outputs the observed particle coordinates at $\mathcal{T}$} following ODE dynamics, with $\mathcal{T}$ being a set of time indices. Starting with an initial set $A_0$, the Neural ODE function $\phi_\theta^{\boldsymbol{v}}$ predicts the subsequent sets of data points over the time indices in $\mathcal{T}$. Specifically, the predicted sets $\widehat{A}_1, \ldots, \widehat{A}_{T-1}$ are obtained by applying the Neural ODE function to $A_0$ and the full range of time indices from $1$ to $T-1$, i.e.,
$
\widehat{A}_1, \ldots, \widehat{A}_{T-1} = \revisenew{\phi_\theta^{\boldsymbol{v}}}\left(A_0, \{1, \ldots, T-1\}\right).
$
Similarly, we define $\revisenew{\phi_\theta^g}: \mathbb{R} \times \mathcal{T} \rightarrow \mathbb{R}^{|\mathcal{T}|}$as the \revisenew{particle weight mapping function of the Neural ODE $\rmd \log w_i(t)/\rmd t=g_\theta(\boldsymbol{x}_i(t), t)$, which inputs a given weight of particle $i$ as the initial condition and outputs the sampled weights at $\mathcal{T}$ following ODE dynamics. For simplicity of notation in algorithm description \cref{algo:uddsb}, we also use $w(A)$ to denote the set of weights of particles in set $A$}. Assume the sample size $N_0$ at the initial time point, with the relative total masses at subsequent time points denoted as $n_{t_k}=N_{k}/N_{0}$ for $k=0,1, \cdots, T-1$. Along the trajectory, each sampled particle $i$ has a weight $w_i(t)$ with the initial condition
$w_{i}(0)=1/N_0$.
We then establish a mapping $h_k$ from points at time $t_k$ to the sampled particles (total number $N_0$) predicted at time $t_k$, which links the real data points in $A_{k}$ to its closest point in sampled particles at $\widehat{A}_{k}$, i.e.,
$
h_k: A_{k}\rightarrow \widehat{A}_{k}, \ h_k(\boldsymbol{{x}}_{t_{k}})=\text{argmin}_{\boldsymbol{y} \in \widehat{A}_{k}} \|\boldsymbol{{x}}_{t_{k}}- \boldsymbol{y}\|_2^2.
$
The mass matching error at time $t_k$ is then defined as
$
M_{k}=\sum_{i=1}^{N_0}\Big\|w_{i}(t_k)-\text{card}\left(h_k^{-1}(\boldsymbol{x}_i(t_k))\right)\frac{1}{N_0} \Big\|_2^2.
$
Here $\text{card}(A)$ denotes the cardinality (i.e., number of elements for the finite set) of set $A$. This error metric ensures the matching of local masses.  \revise{So the $\mathcal{L}_{\text{Mass}}=\sum_{k=1}^{T-1}M_{k}$} Furthermore, when $M_t=0$, we have $
\sum_{i=1}^{N_0}w_i(t_k)=\sum_{i=1}^{N_0}\text{card}\left(h_t^{-1}(\boldsymbol{x}_i(t_k))\right)/N_0 = N_{k}/N_0=n_{t_k}.
$
So the local matching loss also encourages the matching of total mass, since the left-hand side is the unbiased estimator of $\int_{\mathbb{R}^d}p(\boldsymbol{y},t_k)d\boldsymbol{y}$. 
Once we have determined the weights $w_i(t)$, we utilize these weights to perform optimal transport matching of the distributions
$
\revise{\mathcal{L}_{\text{OT}}}:=\sum_{k=1}^{T-1} \mathcal{W}_2\big(\hat{\boldsymbol{w}}^{k}, \boldsymbol{w}(t_k)\big),
$
where $\hat{\boldsymbol{w}}^{k}=(1/N_{k},1/N_{k},...,1/N_{k})$ is the uniform distribution of $A_k$ at time $t$, $\boldsymbol{w}(t_k)=(w_1(t_k),w_2(t_k),...,w_{N_0}(t_k))/\sum_{i=1}^{N_0}w_i(t_k)$ is the predicted weight distribution of sample particles at time $t_k$ and $\mathcal{W}_2$ represents the Wasserstein distance between $A_k$ and $\widehat{A}_k$ with normalized distribution defined by $\hat{\boldsymbol{w}}^{k}$ and $\boldsymbol{w}(t_k)$.

\subsection{\revise{Fokker-Planck Constraint and Two Stage Training}}
In addition to the energy loss defined in  \eqqref{eq:unddsb_cost} and the reconstruction loss defined in \eqqref{eq:unddsb_recon}, it is necessary to incorporate a physics-informed loss (PINN-loss) \citep{PINN} to constrain the relationships among the three neural networks, i.e., the Fokker-Planck constraint \cref{eq:unddsb_fisher_contineq}. Here we utilize a  Gaussian mixture model to estimate the initial distribution (\cref{appen:pinn}), ensuring that it satisfies the initial conditions $p_0$, and the PINN-loss is defined as
\begin{equation}\label{eq:fp}
  \revise{\mathcal{L}_\text{FP}}=\left\| \partial_t p_{\theta}+\nabla_{\boldsymbol{x}} \cdot\left(p_{\theta} \boldsymbol{v}_{\theta}\right) - g_{\theta}p_{\theta}\right\|+ \lambda_w\left\| p_{\theta}(\boldsymbol{x},0)-p_0\right\|,
\end{equation}
where $p_{\theta}=\exp{\frac{2}{\sigma^2}s_{\theta}}$.
Therefore, the total loss function \eqref{eq:total_loss} is defined as the weighted sum of the energy, the reconstruction error and the Fokker-Planck loss
with the loss weights as hyper-parameters. This allows us to develop a neural network algorithm for solving the RUOT problem (\cref{algo:uddsb,appen:learn_alg}).

\revise{
We adopt a two-stage training approach to deal with multiple loss terms and stabilize the training process. For the pre-training stage, initially, we use reconstruction loss only to train $\boldsymbol{v}_{\theta}$ and $g_{\theta}$, ensuring a required matching as the initial value. Subsequently, we fix $\boldsymbol{v}_{\theta}$ and $g_{\theta}$ and employ flow-matching to learn an initial and well-optimized log density function ($s_\theta(\boldsymbol{x}, t)$). Specifically, it is conducted by conditional flow matching \citep{cfm_lipman,cfm_tong,sflowmatch} (Appendix \ref{appen:train_alg}). \revisenew{As part of this pre-training stage, we incorporate a hyperparameter scheduling strategy to further enhance stability (\cref{appen:experdetails}).} Based on these warmup steps in the pre-training stage, in the \revisenew{training stage}, we use the obtained $\boldsymbol{v}_{\theta}$, $g_{\theta}$, and the optimized log density function as the starting point, then get the final result by minimizing the total loss \eqref{eq:total_loss}. 
In summary, by integrating all the methodologies, we derive the  \cref{algo:uddsb} for training the regularized unbalanced optimal transport (\cref{appen:train_alg,appen:initiallog}). We discuss the loss weighting strategy and settings in \cref{appen:experdetails}.}

\begin{algorithm}
\color{black} 
\caption{\revise{Training Regularized Unbalanced Optimal Transport} }
\label{algo:uddsb}
\begin{algorithmic}[1]
\Require Datasets $A_0, \ldots, A_{T-1}$, batch size $N$,  maximum ode iteration $n_{\text{ode}}$, maximum log density iteration $n_{\text{log-density}}$, initialized ODE $\boldsymbol{v_\theta}$, growth  $g_\theta$ and log density $s_\theta$
\Ensure Trained neural ODE $\boldsymbol{v_\theta}$, growth function $g_\theta$ and log density function $s_\theta$.
\State \textbf{Pre-Training Stage:}
\For {$i=1$ to $n_{\text{ode}}$} \Comment{Distribution Reconstruction training}
    \For {$t=0$ to $T-2$}
        \State \revisenew{$\widehat{A}_{t+1} \leftarrow \phi_\theta^{\boldsymbol{v}}\left(\widehat{A}_t, t+1\right)$, 
         {$w(\widehat{A}_{t+1})\leftarrow \phi_\theta^g\left(w(\widehat{A}_t), t+1\right)$}}.
         \State \revisenew{$\mathcal{L}_\text{Recons}  \leftarrow\mathcal{L}_\text{Recons} +\lambda_m M_{t} + \lambda_d \mathcal{W}_2\big(\hat{\boldsymbol{w}}^{t}, \boldsymbol{w}(t)\big)$} \eqref{eq:unddsb_recon},  \text{update} $\boldsymbol{v_\theta}$ and $g_\theta$ \text{w.r.t. } $\mathcal{L}_\text{Recons}$  with \revisenew{hyperparameter scheduling (\cref{appen:experdetails}).}
\EndFor
\EndFor
    \For {$t=0$ to $T-2$}
%    \Comment{Score matching training}
      \State $\widehat{A}_{t+1} \leftarrow \phi_\theta^{\boldsymbol{v}}\left(\widehat{A}_t, t+1\right)$
      \Comment{Generating samples from learned $\boldsymbol{\boldsymbol{v_\theta}}$.}
      \EndFor
\For {$i=1$ to $n_{\text{log-density}}$} 
\Comment{CFM Score matching \citep{sflowmatch}}
    \State $(\boldsymbol{x}_0, \boldsymbol{x}_1) \sim q(\boldsymbol{x}_0, \boldsymbol{x}_1) ; \quad t \sim \mathcal{U}(0,1) ; \quad \boldsymbol{x} \sim p(\boldsymbol{x}, t \mid (\boldsymbol{x}_0, \boldsymbol{x}_1))$ at generated datasets $\widehat{A}_0, \cdots, \widehat{A}_{T-1}$, $\mathcal{L}_\text{score} \leftarrow \left\|\lambda_s(t) \nabla_{\boldsymbol{x}} s_{\theta}(\boldsymbol{x},t)+\boldsymbol{\epsilon_1}\right\|_2^2 \eqref{eq:score}$
, \text{update} $s_\theta$ \text{w.r.t. the loss} $\mathcal{L}_\text{score}$
\EndFor
\State \textbf{\revisenew{Training Stage}}: 
\For {$i=1$ to $n_{\text{ode}}$} 
\State \revisenew{Estimate the initial distribution through Gaussian Mixture Model (\cref{appen:pinn}).}
    \For {$t=0$ to $T-2$}
        \State \revisenew{$\widehat{A}_{t+1} \leftarrow \phi_\theta^{\boldsymbol{v}}\left(\widehat{A}_t, t+1\right)$,  {$w(\widehat{A}_{t+1})\leftarrow \phi_\theta^{g}\left(w(\widehat{A}_t), t+1\right)$}}
        \State \revisenew{$\mathcal{L}_{\text{Energy}} \leftarrow \mathbb{E}_{\boldsymbol{x}_t \sim p_t}\int_t^{t+1}\left[ \frac{1}{2}\left\|\boldsymbol{v}_\theta\right\|_2^2+\frac{1}{2}\left\|\nabla_{\boldsymbol{x}} s_\theta\right\|_2^2-\left(\frac{\sigma^2}{2}+s_\theta\right)g_\theta- \frac{(\sigma^2(t))'}{\sigma^2(t)}s_\theta+ \alpha \Psi\left(g_\theta\right) \right]{w}(z)\mathrm{d} z$}
        \State \revisenew{$\mathcal{L}_\text{Recons} \leftarrow \mathcal{L}_{\text{Recons}}+\lambda_m M_{t} + \lambda_d \mathcal{W}_2\big(\hat{\boldsymbol{w}}^{t}, \boldsymbol{w}(t)\big)$ \eqref{eq:unddsb_recon}}
        \State \revisenew{$\mathcal{L}_\text{FP}\leftarrow\left\| \partial_t p_{\theta}+\nabla_{\boldsymbol{x}} \cdot\left(p_{\theta} \boldsymbol{v}_\theta(\boldsymbol{x},t)\right) - g_{\theta}(\boldsymbol{x},t) p_{\theta}\right\|+ \lambda_w\left\| p_{\theta}(\boldsymbol{x},0)-p_0\right\|$ \eqref{eq:fp}}
    %\EndFor
    \State \revisenew{$\mathcal{L}\leftarrow  \mathcal{L}_{\text{Energy}}+\lambda_r\mathcal{L}_{\text{Recons}}+\lambda_f\mathcal{L}_{\text{FP}}$ \eqref{eq:total_loss}}
     \revisenew{, \text{update} $\boldsymbol{v_\theta}$, $g_\theta$ and $s_\theta$ \text{w.r.t.} $\mathcal{L}$}
\EndFor
\EndFor
\end{algorithmic}
\end{algorithm}

\section{Numerical Results}\label{sec:numerical} \revisenew{This section evaluates the DeepRUOT solver's ability to recover growth and transitions accurately and produce realistic stochastic dynamics for constructing a Waddington’s developmental landscape.}  We take $\Psi\left(\gxt\right)=|\gxt|_2^2$ and $\sigma(t)$ is constant in the following computations.

\begin{figure}[htp]
    \centering
    \includegraphics[width=\linewidth]{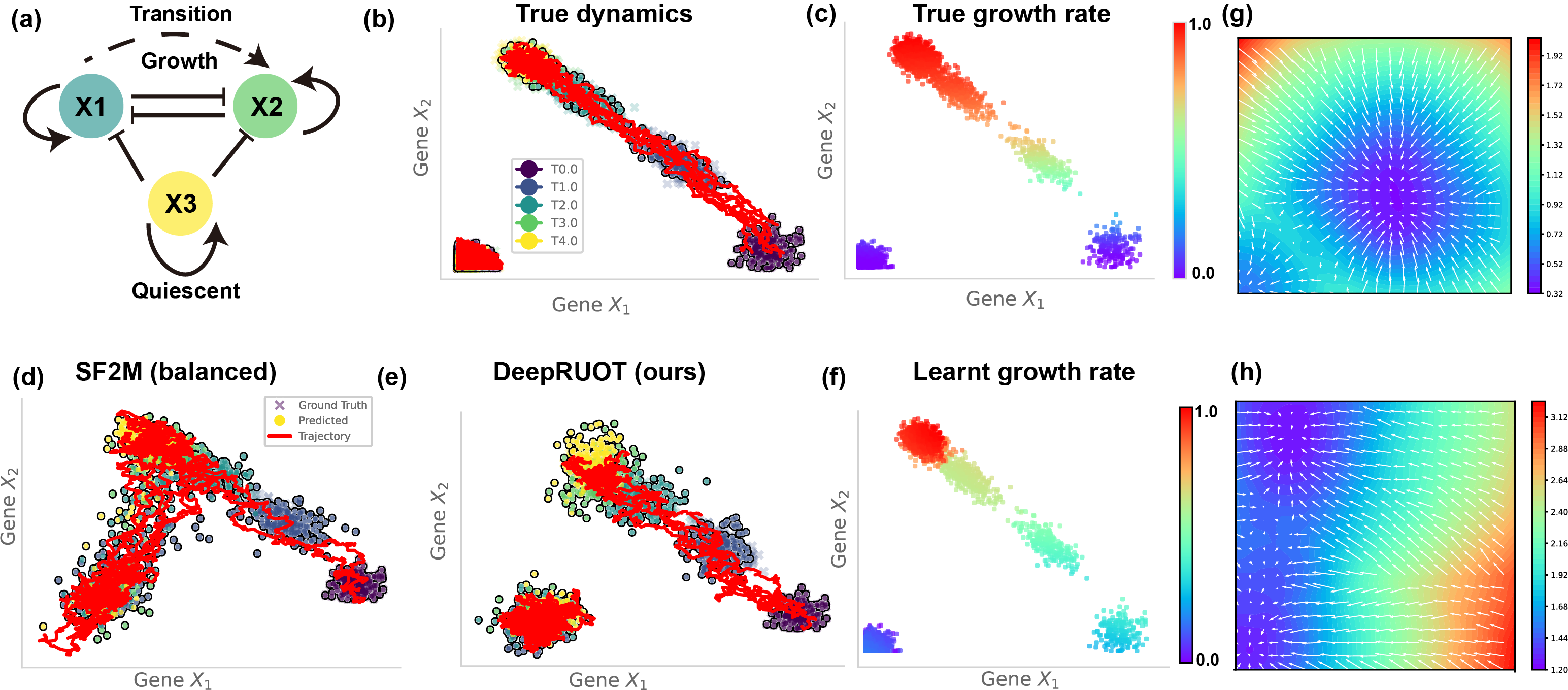}
    \caption{(a) Illustration of the synthetic gene regulatory dynamics. (b) The ground truth  cellular dynamics project on $(X_1 , X_2)$. (c) The ground truth growth rates. (d) The dynamics learned by balanced Schrödinger bridge (SF2M \citep{sflowmatch}, $\sigma=0.25$). (e) The dynamics learned by our DeepRUOT solver ($\sigma=0.25$). (f) The growth rates inferred by our DeepRUOT solver. (g) The Waddington developmental  landscape learned at $t=1$. (h) The constructed landscape at $t=4$. }
    \label{fig:gene_fig1}
\end{figure}

\paragraph{Synthetic Gene Regulatory Network}
Inspired by \citep{Tigon}, we adopt the same three-gene simulation model (Appendix \ref{appen:gene}) to explore the stochastic dynamics of gene regulation, as illustrated in \cref{fig:gene_fig1}(a). The resulting gene regulatory and cellular dynamics are illustrated in  \cref{fig:gene_fig1}(b), (c), where a quiescent region and an area exhibiting both transition and growth can be observed. We focus on the projection of these dynamics onto the two-dimensional space of $(X_1, X_2)$ since on $X_3$ it remains quiescent. In \cref{fig:gene_fig1}(d-e), we compare DeepRUOT with the balanced diffusion Schrödinger bridge method as described in \citep{sflowmatch}. We find that neglecting the growth in the balanced diffusion Schrödinger bridge leads to a false transition, incorrectly attracting cells in the quiescent state to the transition and growth region. 
The underlying reason is that the growth factor causes \revise{an} increase in the number of cells. If the growth factor is ignored, the regions with increasing cell numbers will attract cells to transition into them to maintain balance. Our approach, which explicitly incorporates growth dynamics, indeed eliminates false transitions and yields results consistent with the ground truth dynamics  \cref{fig:gene_fig1}(b). Furthermore, the growth rates inferred by DeepRUOT (\cref{fig:gene_fig1}(f)) closely match the ground truth after normalization, demonstrating the robustness and accuracy of the DeepRUOT solver (\cref{fig:gene_fig1}(c)). 
In \cref{table:gene}, we present quantitative metrics ($\mathcal{W}_1$ and $\mathcal{W}_2$, \cref{appen:evalu}) to evaluate the performance of our proposed algorithm in comparison with several representative baseline methods \revise{(Balanced OT, Balanced SB,  Unbalanced OT, Unbalanced Action Matching (AM), Unbalanced SB)}. The results demonstrate that DeepRUOT is quantitatively more accurate than the competing methods. Furthermore, we observe that our algorithm can benefit from the incorporation of stochasticity, outperforming the solver without diffusion (\revise{Unbalanced OT}). \revise{We include the detail ablation studies in \cref{appen:ablation}.}

\begin{table}[th]
\centering
\caption{Wasserstein distance ($\mathcal{W}_1$ and $\mathcal{W}_2$)  of predictions at different points  across five runs on gene regulatory data (\cref{appen:evalu}). We show the mean value with one standard deviation.}
\begin{threeparttable}
\resizebox{\textwidth}{!}{ % Resize table to fit the width of the text
\begin{tabular}{lcccccccc}
\toprule
& \multicolumn{2}{c}{  $t=1$} & \multicolumn{2}{c}{  $t=2$} & \multicolumn{2}{c}{  $t=3$} & \multicolumn{2}{c}{  $t=4$} \\
\cmidrule(r){2-3} \cmidrule(r){4-5} \cmidrule(r){6-7} \cmidrule(r){8-9}
Model & $\mathcal{W}_1$ & $\mathcal{W}_2$ & $\mathcal{W}_1$ & $\mathcal{W}_2$ & $\mathcal{W}_1$ & $\mathcal{W}_2$ & $\mathcal{W}_1$ & $\mathcal{W}_2$ \\
\midrule
MIOFlow \citep{mioflow} & $0.098_{\pm 0.000}$ & $0.113_{\pm 0.000}$ & $0.250_{\pm 0.000}$ & $0.295_{\pm 0.000}$ & $0.421_{\pm 0.000}$ & $0.536_{\pm 0.000}$ & $0.614_{\pm 0.000}$ & $0.802_{\pm 0.000}$ \\
SF2M \citep{sflowmatch} & $0.174_{\pm 0.010}$ & $0.303_{\pm 0.023}$ & $0.430_{\pm 0.027}$ & $0.719_{\pm 0.032}$ & $0.686_{\pm 0.054}$ & $1.050_{\pm 0.047}$ & $0.871_{\pm 0.072}$ & 
$1.242_{\pm 0.065}$ \\
\revise{Unbalanced SB} \citep{bunne_unsb}  & \revise{$0.729_{\pm 0.009}$} & \revise{$0.846_{\pm 0.012}$} & \revise{${0.747}_{\pm 0.009}$} & \revise{${0.823}_{\pm 0.012}$}& \revise{$0.612_{\pm 0.012}$} & 
\revise{$0.725_{\pm 0.018}$} & 
\revise{${0.572}_{\pm 0.029}$} &
\revise{${0.944}_{\pm 0.031}$}
\\
\revise{uAM} \citep{action_matching}  & \revise{$0.448_{\pm 0.000}$} & \revise{$0.495_{\pm 0.000}$} & \revise{$0.650_{\pm 0.000}$} & \revise{$0.691_{\pm 0.000}
$} & \revise{${0.661}_{\pm 0.000}$} & \revise{${0.749}_{\pm 0.000}$} & \revise{$0.672_{\pm 0.000}$} & \revise{${0.864}_{\pm 0.000}$} \\
\revise{Unbalanced OT} \citep{Tigon}  & $0.047_{\pm 0.000}$ & $0.060_{\pm 0.000}$ & $0.059_{\pm 0.000}$ & $0.088_{\pm 0.000}
$ & ${0.073}_{\pm 0.000}$ & ${0.084}_{\pm 0.000}$ & $0.107_{\pm 0.000}$ & ${0.124}_{\pm 0.000}$ \\
DeepRUOT (ours) & $\boldsymbol{0.044}_{\pm 0.001}$ & $\boldsymbol{0.058}_{\pm 0.001}$ & $\boldsymbol{0.056}_{\pm 0.002}$ & $\boldsymbol{0.084}_{\pm 0.002}$ & $\boldsymbol{0.071}_{\pm 0.002}$ & $\boldsymbol{0.083}_{\pm 0.002}$ & $\boldsymbol{0.104}_{\pm 0.001}$ & $\boldsymbol{0.121}_{\pm 0.000}$ \\
\bottomrule
\end{tabular}
} % End of resizebox

  \end{threeparttable}
  \label{table:gene}
\end{table}

{ \bf Learning Waddington Developmental Landscape} In biophysics, Waddington’s landscape metaphor is a well-known model for representing the cell fate decision process. Constructing such a potential landscape has been widely studied \citep{landscape_PZ,landscape_Shi,epr,lichunhe2013quantifying,Lichunhe2010potential,lichunhe2014landscape,bian2023improved,bian2024quantifying,zhou2024spatial,zhou2024revealing}, however, it still remains a challenging problem in the single-cell omics data. The energy landscape is defined by
$
U=-\sigma^2 \log p_{\text{ss}}/2
$
where $p_{\text{ss}}$ is the steady-state PDF satisfying the Fokker-Planck equation
$
-\nabla \cdot (p_{ss} \boldsymbol{b}) +\frac{\sigma^2}{2} \Delta p_{\text{ss}}= g p_{\text{ss}}
$.
In this work, leveraging our model, we can naturally infer the time-varying potential energy landscape through the learned log density function, and use $-\sigma^2(t)\log p(\boldsymbol{x},t)/2$ to represent the landscape at time $t$. The lower energy function indicates more stable cell fates in the landscape. In \cref{fig:gene_fig1}(g-h), we can observe the temporal evolution of the potential landscape in the synthetic gene regulation system. The quiescent cells always occupy a potential well, indicating its stability. Interestingly, the location of potential wells belonging to the transitional cell population is moving in a consistent direction with ground truth cell-state transition dynamics. Overall, the results suggest the accuracy and usefulness of DeepRUOT algorithm.

\begin{figure}[htp]
    \centering
    \includegraphics[width=0.9\linewidth]{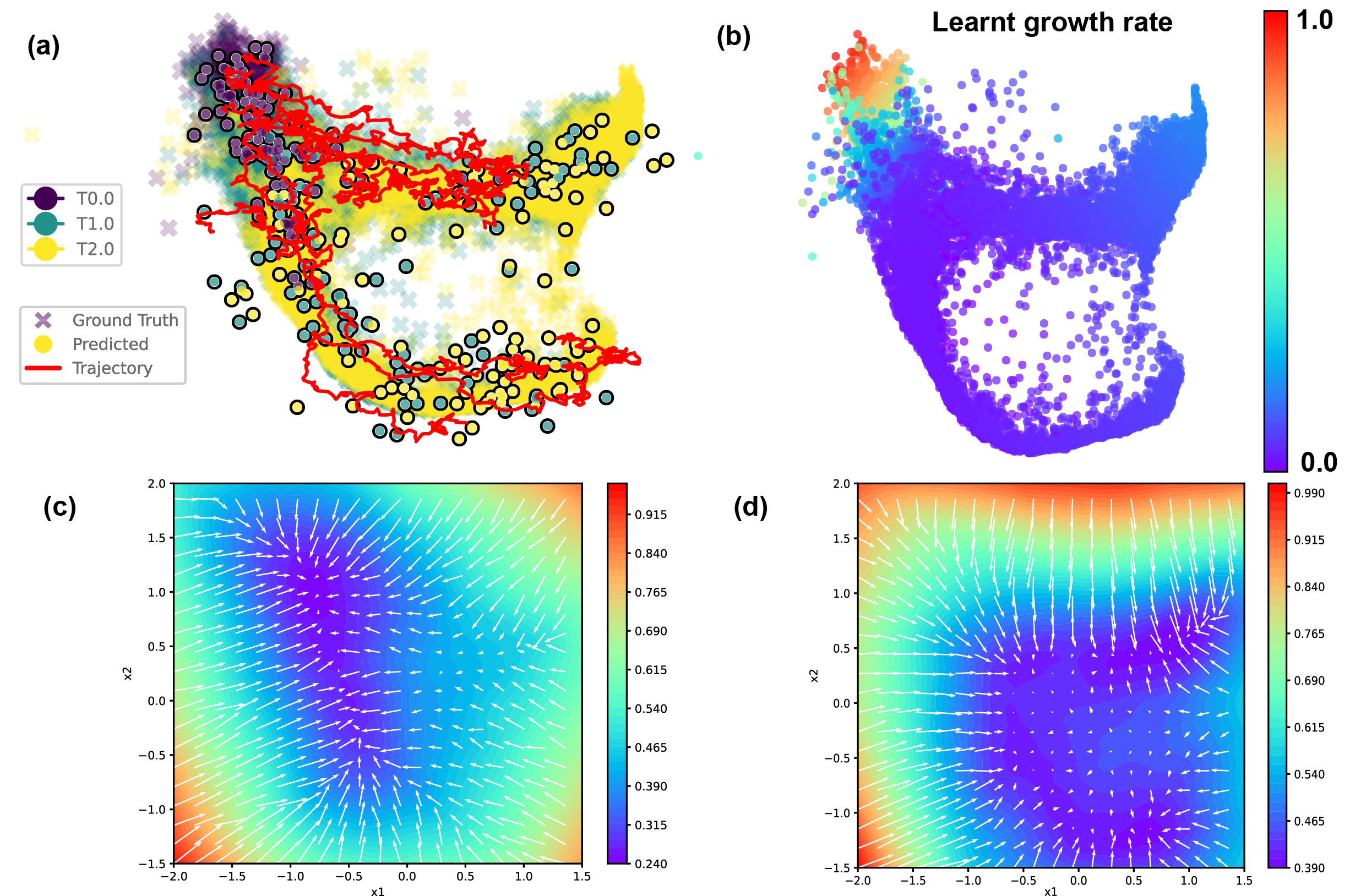}
    \caption{Application in hematopoiesis scRNA-seq data ($\sigma=0.25$). (a) The stochastic dynamics learned by RUOT. (b) The growth rates learned by DeepRUOT. (c) The constructed Waddington developmental landscape at $t=0$. (d) The landscape at $t=2$.}
    \label{fig:scRNA_1}
\end{figure}

\paragraph{Synthetic Gaussian Mixtures}
Inspired by \citep{Liwu2020machine}, we employed a high-dimensional Gaussian mixture model to evaluate the scalability of DeepRUOT. Initially, we generated a 10-dimensional Gaussian mixture distribution. The initial density is represented as a Gaussian mixture derived from the average of two Gaussians, with their projections onto the $(x_1, x_2)$ plane depicted in \cref{appen:gauss}. The final density is also modeled as a mixture of three Gaussians, with its projection onto $(x_1, x_2)$ planes. \cref{fig:gauss_10d} (\cref{appen:gauss}) demonstrates that our model effectively incorporates growth while learning the stochastic dynamics that transition from the initial to the target distribution. We observe that the cells in the upper region exhibit proliferation without transport. If the true cell dynamics indeed follow this pattern, models that neglect the growth factor, such as conventional balanced methods, would fail to capture these dynamics. Conversely, if the true dynamics involve transport from the lower to the upper cells, our model can still recover such dynamics, provided more detailed temporal resolution data is available. This highlights the robustness of our approach in capturing complex dynamical behaviors involving both growth and transport. Similarly, we visualized the energy landscapes at the initial and final time points \cref{fig:gauss_10d} (\cref{appen:gauss}). The results reveal that the cells ultimately differentiate into three distinct fate states.
\paragraph{Real Single-Cell Population Dynamics} Next, we evaluate our algorithm on a real scRNA-seq dataset. We use the same dataset as in \citep{Tigon,weinreb2020lineage}, which involves mouse hematopoiesis analyzed by using a lineage tracing technique. After batch correction across different experiments, the data was projected onto 2D reduced-dimension force-directed layouts (SPRING plots). A clear bifurcation is observed where early-stage progenitor cells differentiate into two distinct fates (\cref{fig:scRNA_1}(a)). Using the RUOT-based approach, we learn the underlying stochastic dynamics of the data, along with the growth rates and developmental landscape at different times (\cref{fig:scRNA_1}(b),(c),(d)). 
We similarly evaluated quantitative metrics to compare our method with several baseline approaches (\cref{table:scRNA}). We find that our method outperforms others (\cref{appen:sc-mouse}). \revise{We then test our method on a time-series single cell dataset from an A549 cancer cell line, where cells were exposed to TGFB1 to induce EM \citep{Tigon}. We project the data with PCA on a ten-dimensional latent space as the input of our algorithm. The results are shown in \cref{appen:scRNA-emt}, which indicates that DeepRUOT remains effective and applicable in higher-dimensional settings, demonstrating its versatility and robustness in modeling complex single-cell dynamics.} 
\begin{table}[th]
\centering
\caption{Wasserstein distance ($\mathcal{W}_1$ and $\mathcal{W}_2$)  of predictions at different points  across five runs on \revise{scRNA-seq data}        (\cref{appen:evalu}, $\sigma=0.25$). We show the mean value with one standard deviation.}
\begin{threeparttable}
\resizebox{0.8\textwidth}{!}{ % 将表格宽度调整为文本宽度的 80%
\begin{tabular}{lcccc}
\toprule
& \multicolumn{2}{c}{  $t=1$} & \multicolumn{2}{c}{  $t=2$} \\
\cmidrule(r){2-3} \cmidrule(r){4-5}
Model & $\mathcal{W}_1$ & $\mathcal{W}_2$ & $\mathcal{W}_1$ & $\mathcal{W}_2$ \\
\midrule
MIOFlow \citep{mioflow} & $0.276725_{\pm 0.000}$ & $0.312102_{\pm 0.000}$ & $0.307610_{\pm 0.000}$ & $	0.402190_{\pm 0.000}$ \\
SF2M \citep{sflowmatch} & $0.167477_{\pm 0.003}$ & $0.213489_{\pm 0.006}$ & ${0.190020}_{\pm 0.016}$ & ${0.241516}_{\pm 0.022}$ \\
\revise{Unbalanced SB} \citep{bunne_unsb}  & \revise{$0.387538_{\pm 0.009}$} & \revise{$0.460603_{\pm 0.008}$} & \revise{$\boldsymbol{0.128254}_{\pm 0.003}$} & \revise{${0.188339	}_{\pm 0.009}$}\\
\revise{uAM \citep{action_matching}}  & \revise{$0.744918_{\pm 0.000}$} & \revise{$0.851704_{\pm 0.000}$} & \revise{${0.777237}_{\pm 0.000}$} & \revise{${0.889527	}_{\pm 0.000}$}\\
\revise{Unbalanced OT} \citep{Tigon}  & $0.313522_{\pm 0.000}$ & $0.396947_{\pm 0.000}$ & $0.342230_{\pm 0.000}$ & $0.469342_{\pm 0.000}$ \\
DeepRUOT (ours) & $\boldsymbol{0.145026}_{\pm 0.002}$ & $\boldsymbol{0.172878}_{\pm 	0.002}$ & $0.132411_{\pm 0.006}$ & $\boldsymbol{0.167328}_{\pm 0.010}$ \\
\bottomrule
\end{tabular}
}% 结束 resizebox
  \end{threeparttable}
  \label{table:scRNA}
\end{table}

\section{Conclusion}\label{sec:conclu}
We have introduced DeepRUOT for learning regularized unbalanced optimal transport (RUOT) and continuous unbalanced stochastic dynamics from time-series snapshot data. By leveraging Fisher regularization, our method transforms an SDE problem into an ODE constraint. Through the use of neural network modeling for growth and death, our framework models dynamics without requiring prior knowledge of these processes. We have demonstrated the effectiveness of our method on a synthetic gene regulatory network, high-dimensional Gaussian Mixture Model, and single-cell RNA-seq data, showing its ability to eliminate false transitions caused by neglecting growth processes.
Future directions involve extending the methodology to learn the cell latent embedding space jointly with the dynamics and developing more computationally efficient algorithms. We expect future work to address these aspects and improve trajectory inference problems in various scenarios, including multi-omics data and other scenarios beyond biology in machine learning.

\section*{Acknowledgments} 
We thank Dr. Hugo Lavenant (Bocconi) for helpful discussions on Section E.1 and Qiangwei Peng (PKU) for helpful suggestions on the algorithm. We thank Prof. Qing Nie (UCI), Prof. Chunhe Li (Fudan), and Juntan Liu (Fudan) for their insightful discussions. This work was supported by the National Key R\&D Program of China (No. 2021YFA1003301 to T.L.) and 
 National Natural Science Foundation of China (NSFC No. 12288101 to T.L. \& P.Z., and 8206100646, T2321001 to P.Z.).  We acknowledge the support from the High-performance Computing Platform of Peking University for computation.  We thank the anonymous referees for their valuable feedback and constructive suggestions.

\bibliography{iclr2025_conference}

\begin{thebibliography}{97}
\providecommand{\natexlab}[1]{#1}
\providecommand{\url}[1]{\texttt{#1}}
\expandafter\ifx\csname urlstyle\endcsname\relax
  \providecommand{\doi}[1]{doi: #1}\else
  \providecommand{\doi}{doi: \begingroup \urlstyle{rm}\Url}\fi

\bibitem[Albergo et~al.(2023)Albergo, Boffi, and Vanden-Eijnden]{albergo2023stochastic}
Michael~S Albergo, Nicholas~M Boffi, and Eric Vanden-Eijnden.
\newblock Stochastic interpolants: A unifying framework for flows and diffusions.
\newblock \emph{arXiv preprint arXiv:2303.08797}, 2023.

\bibitem[Albergo \& Vanden-Eijnden(2023)Albergo and Vanden-Eijnden]{albergo2023building}
Michael~Samuel Albergo and Eric Vanden-Eijnden.
\newblock Building normalizing flows with stochastic interpolants.
\newblock In \emph{The Eleventh International Conference on Learning Representations}, 2023.

\bibitem[Baradat \& Lavenant(2021)Baradat and Lavenant]{branRUOT}
Aymeric Baradat and Hugo Lavenant.
\newblock Regularized unbalanced optimal transport as entropy minimization with respect to branching brownian motion.
\newblock \emph{arXiv preprint arXiv:2111.01666}, 2021.

\bibitem[Benamou \& Brenier(2000)Benamou and Brenier]{benamou2000computational}
Jean-David Benamou and Yann Brenier.
\newblock A computational fluid mechanics solution to the monge-kantorovich mass transfer problem.
\newblock \emph{Numerische Mathematik}, 84\penalty0 (3):\penalty0 375--393, 2000.

\bibitem[Bian et~al.(2023)Bian, Zhang, and Li]{bian2023improved}
Shirui Bian, Yunxin Zhang, and Chunhe Li.
\newblock An improved approach for calculating energy landscape of gene networks from moment equations.
\newblock \emph{Chaos: An Interdisciplinary Journal of Nonlinear Science}, 33\penalty0 (2), 2023.

\bibitem[Bian et~al.(2024)Bian, Zhou, Lin, and Li]{bian2024quantifying}
Shirui Bian, Ruisong Zhou, Wei Lin, and Chunhe Li.
\newblock Quantifying energy landscape of oscillatory systems: Explosion, pre-solution, and diffusion decomposition.
\newblock \emph{arXiv preprint arXiv:2401.06959}, 2024.

\bibitem[Bunne et~al.(2023{\natexlab{a}})Bunne, Hsieh, Cuturi, and Krause]{bunne_dynam_SB}
Charlotte Bunne, Ya-Ping Hsieh, Marco Cuturi, and Andreas Krause.
\newblock The schr{\"o}dinger bridge between gaussian measures has a closed form.
\newblock In \emph{International Conference on Artificial Intelligence and Statistics}, pp.\  5802--5833. PMLR, 2023{\natexlab{a}}.

\bibitem[Bunne et~al.(2023{\natexlab{b}})Bunne, Stark, Gut, Del~Castillo, Levesque, Lehmann, Pelkmans, Krause, and R{\"a}tsch]{bunne2023learning}
Charlotte Bunne, Stefan~G Stark, Gabriele Gut, Jacobo~Sarabia Del~Castillo, Mitch Levesque, Kjong-Van Lehmann, Lucas Pelkmans, Andreas Krause, and Gunnar R{\"a}tsch.
\newblock Learning single-cell perturbation responses using neural optimal transport.
\newblock \emph{Nature methods}, 20\penalty0 (11):\penalty0 1759--1768, 2023{\natexlab{b}}.

\bibitem[Bunne et~al.(2024)Bunne, Schiebinger, Krause, Regev, and Cuturi]{bunne2024optimal}
Charlotte Bunne, Geoffrey Schiebinger, Andreas Krause, Aviv Regev, and Marco Cuturi.
\newblock Optimal transport for single-cell and spatial omics.
\newblock \emph{Nature Reviews Methods Primers}, 4\penalty0 (1):\penalty0 58, 2024.

\bibitem[Buze \& Duong(2023)Buze and Duong]{highorderRUOT}
Maciej Buze and Manh~Hong Duong.
\newblock Entropic regularisation of unbalanced optimal transportation problems.
\newblock \emph{arXiv preprint arXiv:2305.02410}, 2023.

\bibitem[Cang \& Zhao(2024)Cang and Zhao]{cang2024synchronized}
Zixuan Cang and Yanxiang Zhao.
\newblock Synchronized optimal transport for joint modeling of dynamics across multiple spaces.
\newblock \emph{arXiv preprint arXiv:2406.03319}, 2024.

\bibitem[Cao et~al.(2022)Cao, Gong, Hong, and Wan]{cao2022unified}
Kai Cao, Qiyu Gong, Yiguang Hong, and Lin Wan.
\newblock A unified computational framework for single-cell data integration with optimal transport.
\newblock \emph{Nature Communications}, 13\penalty0 (1):\penalty0 7419, 2022.

\bibitem[Chen et~al.(2018)Chen, Rubanova, Bettencourt, and Duvenaud]{chen2018neuralode}
Ricky T.~Q. Chen, Yulia Rubanova, Jesse Bettencourt, and David Duvenaud.
\newblock Neural ordinary differential equations.
\newblock \emph{Advances in Neural Information Processing Systems}, 2018.

\bibitem[Chen et~al.(2022{\natexlab{a}})Chen, Liu, and Theodorou]{chen2022likelihood}
Tianrong Chen, Guan-Horng Liu, and Evangelos Theodorou.
\newblock Likelihood training of schr\"odinger bridge using forward-backward {SDE}s theory.
\newblock In \emph{International Conference on Learning Representations}, 2022{\natexlab{a}}.

\bibitem[Chen et~al.(2016)Chen, Georgiou, and Pavon]{sb_chen}
Yongxin Chen, Tryphon~T Georgiou, and Michele Pavon.
\newblock On the relation between optimal transport and schr{\"o}dinger bridges: A stochastic control viewpoint.
\newblock \emph{Journal of Optimization Theory and Applications}, 169:\penalty0 671--691, 2016.

\bibitem[Chen et~al.(2022{\natexlab{b}})Chen, Georgiou, and Pavon]{chen2022most}
Yongxin Chen, Tryphon~T Georgiou, and Michele Pavon.
\newblock The most likely evolution of diffusing and vanishing particles: Schrodinger bridges with unbalanced marginals.
\newblock \emph{SIAM Journal on Control and Optimization}, 60\penalty0 (4):\penalty0 2016--2039, 2022{\natexlab{b}}.

\bibitem[Cheng et~al.(2024{\natexlab{a}})Cheng, Liu, Chen, and Shen]{shenjie2024new}
Qing Cheng, Qianqian Liu, Wenbin Chen, and Jie Shen.
\newblock A new flow dynamic approach for wasserstein gradient flows.
\newblock \emph{arXiv preprint arXiv:2406.14870}, 2024{\natexlab{a}}.

\bibitem[Cheng et~al.(2024{\natexlab{b}})Cheng, Lu, Tan, and Xie]{Lujianfeng2024convergence}
Xiuyuan Cheng, Jianfeng Lu, Yixin Tan, and Yao Xie.
\newblock Convergence of flow-based generative models via proximal gradient descent in wasserstein space.
\newblock \emph{IEEE Transactions on Information Theory}, 2024{\natexlab{b}}.

\bibitem[Chizat et~al.(2018{\natexlab{a}})Chizat, Peyr{\'e}, Schmitzer, and Vialard]{uot1}
Lenaic Chizat, Gabriel Peyr{\'e}, Bernhard Schmitzer, and Fran{\c{c}}ois-Xavier Vialard.
\newblock An interpolating distance between optimal transport and fisher--rao metrics.
\newblock \emph{Foundations of Computational Mathematics}, 18:\penalty0 1--44, 2018{\natexlab{a}}.

\bibitem[Chizat et~al.(2018{\natexlab{b}})Chizat, Peyr{\'e}, Schmitzer, and Vialard]{uot2}
Lenaic Chizat, Gabriel Peyr{\'e}, Bernhard Schmitzer, and Fran{\c{c}}ois-Xavier Vialard.
\newblock Unbalanced optimal transport: Dynamic and kantorovich formulations.
\newblock \emph{Journal of Functional Analysis}, 274\penalty0 (11):\penalty0 3090--3123, 2018{\natexlab{b}}.

\bibitem[Chizat et~al.(2022)Chizat, Zhang, Heitz, and Schiebinger]{Geofftrajectory2}
L{\'e}na{\"\i}c Chizat, Stephen Zhang, Matthieu Heitz, and Geoffrey Schiebinger.
\newblock Trajectory inference via mean-field langevin in path space.
\newblock \emph{Advances in Neural Information Processing Systems}, 35:\penalty0 16731--16742, 2022.

\bibitem[Chow et~al.(2020)Chow, Li, and Zhou]{LiwuZHou2020wasserstein}
Shui-Nee Chow, Wuchen Li, and Haomin Zhou.
\newblock Wasserstein hamiltonian flows.
\newblock \emph{Journal of Differential Equations}, 268\penalty0 (3):\penalty0 1205--1219, 2020.

\bibitem[Dai~Pra(1991)]{dai1991stochastic}
Paolo Dai~Pra.
\newblock A stochastic control approach to reciprocal diffusion processes.
\newblock \emph{Applied mathematics and Optimization}, 23\penalty0 (1):\penalty0 313--329, 1991.

\bibitem[De~Bortoli et~al.(2021)De~Bortoli, Thornton, Heng, and Doucet]{de2021diffusion}
Valentin De~Bortoli, James Thornton, Jeremy Heng, and Arnaud Doucet.
\newblock Diffusion schr{\"o}dinger bridge with applications to score-based generative modeling.
\newblock \emph{Advances in Neural Information Processing Systems}, 34:\penalty0 17695--17709, 2021.

\bibitem[Demetci et~al.(2022)Demetci, Santorella, Sandstede, Noble, and Singh]{demetci2022scot}
Pinar Demetci, Rebecca Santorella, Bj{\"o}rn Sandstede, William~Stafford Noble, and Ritambhara Singh.
\newblock Scot: single-cell multi-omics alignment with optimal transport.
\newblock \emph{Journal of computational biology}, 29\penalty0 (1):\penalty0 3--18, 2022.

\bibitem[Elowitz et~al.(2002)Elowitz, Levine, Siggia, and Swain]{elowitz2002stochastic}
Michael~B Elowitz, Arnold~J Levine, Eric~D Siggia, and Peter~S Swain.
\newblock Stochastic gene expression in a single cell.
\newblock \emph{Science}, 297\penalty0 (5584):\penalty0 1183--1186, 2002.

\bibitem[Flamary et~al.(2021)Flamary, Courty, Gramfort, Alaya, Boisbunon, Chambon, Chapel, Corenflos, Fatras, Fournier, Gautheron, Gayraud, Janati, Rakotomamonjy, Redko, Rolet, Schutz, Seguy, Sutherland, Tavenard, Tong, and Vayer]{flamary2021pot}
R{\'e}mi Flamary, Nicolas Courty, Alexandre Gramfort, Mokhtar~Z. Alaya, Aur{\'e}lie Boisbunon, Stanislas Chambon, Laetitia Chapel, Adrien Corenflos, Kilian Fatras, Nemo Fournier, L{\'e}o Gautheron, Nathalie~T.H. Gayraud, Hicham Janati, Alain Rakotomamonjy, Ievgen Redko, Antoine Rolet, Antony Schutz, Vivien Seguy, Danica~J. Sutherland, Romain Tavenard, Alexander Tong, and Titouan Vayer.
\newblock Pot: Python optimal transport.
\newblock \emph{Journal of Machine Learning Research}, 22\penalty0 (78):\penalty0 1--8, 2021.

\bibitem[Gangbo et~al.(2019)Gangbo, Li, Osher, and Puthawala]{liunnormalized}
Wilfrid Gangbo, Wuchen Li, Stanley Osher, and Michael Puthawala.
\newblock Unnormalized optimal transport.
\newblock \emph{Journal of Computational Physics}, 399:\penalty0 108940, 2019.

\bibitem[Gao et~al.(2024{\natexlab{a}})Gao, Huang, Jiao, and Zheng]{jiao2024convergencecontinuousnormalizingflows}
Yuan Gao, Jian Huang, Yuling Jiao, and Shurong Zheng.
\newblock Convergence of continuous normalizing flows for learning probability distributions.
\newblock \emph{arXiv preprint arXiv:2404.00551}, 2024{\natexlab{a}}.

\bibitem[Gao et~al.(2024{\natexlab{b}})Gao, Cao, and Wan]{gao2024graspot}
Zizhan Gao, Kai Cao, and Lin Wan.
\newblock Graspot: A graph attention network for spatial transcriptomics data integration with optimal transport.
\newblock \emph{bioRxiv}, pp.\  2024--02, 2024{\natexlab{b}}.

\bibitem[Gentil et~al.(2017)Gentil, L{\'e}onard, and Ripani]{sb_Gentil}
Ivan Gentil, Christian L{\'e}onard, and Luigia Ripani.
\newblock About the analogy between optimal transport and minimal entropy.
\newblock In \emph{Annales de la Facult{\'e} des sciences de Toulouse: Math{\'e}matiques}, volume~26, pp.\  569--600, 2017.

\bibitem[Gu et~al.(2024)Gu, Chien, and Greenewald]{gu2024partially}
Anming Gu, Edward Chien, and Kristjan Greenewald.
\newblock Partially observed trajectory inference using optimal transport and a dynamics prior.
\newblock \emph{arXiv preprint arXiv:2406.07475}, 2024.

\bibitem[Ho et~al.(2020)Ho, Jain, and Abbeel]{ho2020denoising}
Jonathan Ho, Ajay Jain, and Pieter Abbeel.
\newblock Denoising diffusion probabilistic models.
\newblock \emph{Advances in neural information processing systems}, 33:\penalty0 6840--6851, 2020.

\bibitem[Huguet et~al.(2022)Huguet, Magruder, Tong, Fasina, Kuchroo, Wolf, and Krishnaswamy]{mioflow}
Guillaume Huguet, Daniel~Sumner Magruder, Alexander Tong, Oluwadamilola Fasina, Manik Kuchroo, Guy Wolf, and Smita Krishnaswamy.
\newblock Manifold interpolating optimal-transport flows for trajectory inference.
\newblock \emph{Advances in neural information processing systems}, 35:\penalty0 29705--29718, 2022.

\bibitem[Janati et~al.(2020)Janati, Muzellec, Peyr{\'e}, and Cuturi]{janati2020entropic}
Hicham Janati, Boris Muzellec, Gabriel Peyr{\'e}, and Marco Cuturi.
\newblock Entropic optimal transport between unbalanced gaussian measures has a closed form.
\newblock \emph{Advances in neural information processing systems}, 33:\penalty0 10468--10479, 2020.

\bibitem[Jiang \& Wan(2024)Jiang and Wan]{jiang2024physics}
Qi~Jiang and Lin Wan.
\newblock {A physics-informed neural SDE network for learning cellular dynamics from time-series scRNA-seq data}.
\newblock \emph{Bioinformatics}, 40:\penalty0 ii120--ii127, 09 2024.
\newblock ISSN 1367-4811.

\bibitem[Jiang et~al.(2022)Jiang, Zhang, and Wan]{jiang2022dynamic}
Qi~Jiang, Shuo Zhang, and Lin Wan.
\newblock Dynamic inference of cell developmental complex energy landscape from time series single-cell transcriptomic data.
\newblock \emph{PLoS Computational Biology}, 18\penalty0 (1):\penalty0 e1009821, 2022.

\bibitem[Jiao et~al.(2024{\natexlab{a}})Jiao, Kang, Lin, Liu, and Zuo]{Jiao2024LatentSB}
Yuling Jiao, Lican Kang, Huazhen Lin, Jin Liu, and Heng Zuo.
\newblock Latent schr{\"o}dinger bridge diffusion model for generative learning.
\newblock \emph{arXiv preprint arXiv:2404.13309}, 2024{\natexlab{a}}.

\bibitem[Jiao et~al.(2024{\natexlab{b}})Jiao, Lai, Wang, and Yan]{jiao2024convergenceanalysisflowmatching}
Yuling Jiao, Yanming Lai, Yang Wang, and Bokai Yan.
\newblock Convergence analysis of flow matching in latent space with transformers.
\newblock \emph{arXiv preprint arXiv:2404.02538}, 2024{\natexlab{b}}.

\bibitem[Jin et~al.(2024)Jin, Liu, Wu, Ye, and Zhou]{zhouhao2024parameterized_W}
Yijie Jin, Shu Liu, Hao Wu, Xiaojing Ye, and Haomin Zhou.
\newblock Parameterized wasserstein gradient flow.
\newblock \emph{arXiv preprint arXiv:2404.19133}, 2024.

\bibitem[Kingma \& Welling(2013)Kingma and Welling]{kingma2022autoencodingvariationalbayes}
Diederik~P Kingma and Max Welling.
\newblock Auto-encoding variational bayes.
\newblock \emph{arXiv preprint arXiv:1312.6114}, 2013.

\bibitem[Klein et~al.(2023{\natexlab{a}})Klein, Palla, Lange, Klein, Piran, Gander, Meng-Papaxanthos, Sterr, Bastidas-Ponce, Tarquis-Medina, et~al.]{moscot}
Dominik Klein, Giovanni Palla, Marius Lange, Michal Klein, Zoe Piran, Manuel Gander, Laetitia Meng-Papaxanthos, Michael Sterr, Aim{\'e}e Bastidas-Ponce, Marta Tarquis-Medina, et~al.
\newblock Mapping cells through time and space with moscot.
\newblock \emph{bioRxiv}, pp.\  2023--05, 2023{\natexlab{a}}.

\bibitem[Klein et~al.(2023{\natexlab{b}})Klein, Uscidda, Theis, and Cuturi]{genot}
Dominik Klein, Th{\'e}o Uscidda, Fabian Theis, and Marco Cuturi.
\newblock Generative entropic neural optimal transport to map within and across spaces.
\newblock \emph{arXiv preprint arXiv:2310.09254}, 2023{\natexlab{b}}.

\bibitem[Koshizuka \& Sato(2023)Koshizuka and Sato]{dyn_sb_koshizuka2023neural}
Takeshi Koshizuka and Issei Sato.
\newblock Neural lagrangian schr\"odinger bridge: Diffusion modeling for population dynamics.
\newblock In \emph{The Eleventh International Conference on Learning Representations}, 2023.

\bibitem[Lahat et~al.(2015)Lahat, Adali, and Jutten]{lahat2015multimodal}
Dana Lahat, T{\"u}lay Adali, and Christian Jutten.
\newblock Multimodal data fusion: an overview of methods, challenges, and prospects.
\newblock \emph{Proceedings of the IEEE}, 103\penalty0 (9):\penalty0 1449--1477, 2015.

\bibitem[Lavenant et~al.(2024)Lavenant, Zhang, Kim, Schiebinger, et~al.]{trajectory}
Hugo Lavenant, Stephen Zhang, Young-Heon Kim, Geoffrey Schiebinger, et~al.
\newblock Toward a mathematical theory of trajectory inference.
\newblock \emph{The Annals of Applied Probability}, 34\penalty0 (1A):\penalty0 428--500, 2024.

\bibitem[L{\'e}ger \& Li(2021)L{\'e}ger and Li]{LiWU2021schb}
Flavien L{\'e}ger and Wuchen Li.
\newblock Hopf--cole transformation via generalized schr{\"o}dinger bridge problem.
\newblock \emph{Journal of Differential Equations}, 274:\penalty0 788--827, 2021.

\bibitem[L{\'e}onard(2014)]{sb}
Christian L{\'e}onard.
\newblock A survey of the schr{\"o}dinger problem and some of its connections with optimal transport.
\newblock \emph{Discrete and Continuous Dynamical Systems-Series A}, 34\penalty0 (4):\penalty0 1533--1574, 2014.

\bibitem[Li \& Wang(2013)Li and Wang]{lichunhe2013quantifying}
Chunhe Li and Jin Wang.
\newblock Quantifying cell fate decisions for differentiation and reprogramming of a human stem cell network: landscape and biological paths.
\newblock \emph{PLoS computational biology}, 9\penalty0 (8):\penalty0 e1003165, 2013.

\bibitem[Li \& Wang(2014)Li and Wang]{lichunhe2014landscape}
Chunhe Li and Jin Wang.
\newblock Landscape and flux reveal a new global view and physical quantification of mammalian cell cycle.
\newblock \emph{Proceedings of the National Academy of Sciences}, 111\penalty0 (39):\penalty0 14130--14135, 2014.

\bibitem[Li et~al.(2020)Li, Lu, and Wang]{li2020fisher}
Wuchen Li, Jianfeng Lu, and Li~Wang.
\newblock Fisher information regularization schemes for wasserstein gradient flows.
\newblock \emph{Journal of Computational Physics}, 416:\penalty0 109449, 2020.

\bibitem[Lipman et~al.(2023)Lipman, Chen, Ben-Hamu, Nickel, and Le]{cfm_lipman}
Yaron Lipman, Ricky T.~Q. Chen, Heli Ben-Hamu, Maximilian Nickel, and Matthew Le.
\newblock Flow matching for generative modeling.
\newblock In \emph{The Eleventh International Conference on Learning Representations}, 2023.

\bibitem[Liu et~al.(2022{\natexlab{a}})Liu, Chen, So, and Theodorou]{Liudeep_genera_sb}
Guan-Horng Liu, Tianrong Chen, Oswin So, and Evangelos Theodorou.
\newblock Deep generalized schr\"{o}dinger bridge.
\newblock In \emph{Advances in Neural Information Processing Systems}, volume~35, pp.\  9374--9388, 2022{\natexlab{a}}.

\bibitem[Liu et~al.(2023{\natexlab{a}})Liu, Vahdat, Huang, Theodorou, Nie, and Anandkumar]{imagesb}
Guan-Horng Liu, Arash Vahdat, De-An Huang, Evangelos~A Theodorou, Weili Nie, and Anima Anandkumar.
\newblock I$^{2}$sb: Image-to-image schr{\"o}dinger bridge.
\newblock \emph{arXiv preprint arXiv:2302.05872}, 2023{\natexlab{a}}.

\bibitem[Liu et~al.(2021)Liu, Ma, Chen, Zha, and Zhou]{zhouhaolearning}
Shu Liu, Shaojun Ma, Yongxin Chen, Hongyuan Zha, and Haomin Zhou.
\newblock Learning high dimensional wasserstein geodesics.
\newblock \emph{arXiv preprint arXiv:2102.02992}, 2021.

\bibitem[Liu et~al.(2022{\natexlab{b}})Liu, Li, Zha, and Zhou]{zhouhao2022neural}
Shu Liu, Wuchen Li, Hongyuan Zha, and Haomin Zhou.
\newblock Neural parametric fokker--planck equation.
\newblock \emph{SIAM Journal on Numerical Analysis}, 60\penalty0 (3):\penalty0 1385--1449, 2022{\natexlab{b}}.

\bibitem[Liu et~al.(2023{\natexlab{b}})Liu, Gong, and Liu]{liu2023flow}
Xingchao Liu, Chengyue Gong, and Qiang Liu.
\newblock Flow straight and fast: Learning to generate and transfer data with rectified flow.
\newblock In \emph{The Eleventh International Conference on Learning Representations}, 2023{\natexlab{b}}.

\bibitem[Maddu et~al.(2024)Maddu, Chard{\`e}s, Shelley, et~al.]{maddu2024inferring}
Suryanarayana Maddu, Victor Chard{\`e}s, Michael Shelley, et~al.
\newblock Inferring biological processes with intrinsic noise from cross-sectional data.
\newblock \emph{arXiv preprint arXiv:2410.07501}, 2024.

\bibitem[Meng et~al.(2024)Meng, Zou, Darbon, and Karniadakis]{meng2024hj}
Tingwei Meng, Zongren Zou, J{\'e}r{\^o}me Darbon, and George~Em Karniadakis.
\newblock Hj-sampler: A bayesian sampler for inverse problems of a stochastic process by leveraging hamilton-jacobi pdes and score-based generative models.
\newblock \emph{arXiv preprint arXiv:2409.09614}, 2024.

\bibitem[Mikami \& Thieullen(2008)Mikami and Thieullen]{rot}
Toshio Mikami and Michele Thieullen.
\newblock Optimal transportation problem by stochastic optimal control.
\newblock \emph{SIAM Journal on Control and Optimization}, 47\penalty0 (3):\penalty0 1127--1139, 2008.

\bibitem[Neklyudov et~al.(2023)Neklyudov, Brekelmans, Severo, and Makhzani]{action_matching}
Kirill Neklyudov, Rob Brekelmans, Daniel Severo, and Alireza Makhzani.
\newblock Action matching: Learning stochastic dynamics from samples.
\newblock In \emph{International conference on machine learning}, pp.\  25858--25889. PMLR, 2023.

\bibitem[Neklyudov et~al.(2024)Neklyudov, Brekelmans, Tong, Atanackovic, Liu, and Makhzani]{tong_action}
Kirill Neklyudov, Rob Brekelmans, Alexander Tong, Lazar Atanackovic, Qiang Liu, and Alireza Makhzani.
\newblock A computational framework for solving wasserstein lagrangian flows.
\newblock In \emph{Forty-first International Conference on Machine Learning}, 2024.

\bibitem[Pariset et~al.(2023)Pariset, Hsieh, Bunne, Krause, and Bortoli]{bunne_unsb}
Matteo Pariset, Ya-Ping Hsieh, Charlotte Bunne, Andreas Krause, and Valentin~De Bortoli.
\newblock Unbalanced diffusion schr\"odinger bridge.
\newblock In \emph{ICML Workshop on New Frontiers in Learning, Control, and Dynamical Systems}, 2023.

\bibitem[Peng et~al.(2024)Peng, Zhou, and Li]{peng2024stvcr}
Qiangwei Peng, Peijie Zhou, and Tiejun Li.
\newblock stvcr: Reconstructing spatio-temporal dynamics of cell development using optimal transport.
\newblock \emph{bioRxiv}, pp.\  2024--06, 2024.

\bibitem[Peyr{\'e} et~al.(2019)Peyr{\'e}, Cuturi, et~al.]{ot_theory}
Gabriel Peyr{\'e}, Marco Cuturi, et~al.
\newblock Computational optimal transport: With applications to data science.
\newblock \emph{Foundations and Trends{\textregistered} in Machine Learning}, 11\penalty0 (5-6):\penalty0 355--607, 2019.

\bibitem[Pooladian \& Niles-Weed(2024)Pooladian and Niles-Weed]{pooladian2024plug}
Aram-Alexandre Pooladian and Jonathan Niles-Weed.
\newblock Plug-in estimation of schr{\"o}dinger bridges.
\newblock \emph{arXiv preprint arXiv:2408.11686}, 2024.

\bibitem[Pooladian et~al.(2024)Pooladian, Domingo-Enrich, Chen, and Amos]{pooladian2024neural}
Aram-Alexandre Pooladian, Carles Domingo-Enrich, Ricky T.~Q. Chen, and Brandon Amos.
\newblock Neural optimal transport with lagrangian costs.
\newblock In \emph{The 40th Conference on Uncertainty in Artificial Intelligence}, 2024.

\bibitem[Raissi et~al.(2019)Raissi, Perdikaris, and Karniadakis]{PINN}
Maziar Raissi, Paris Perdikaris, and George~E Karniadakis.
\newblock Physics-informed neural networks: A deep learning framework for solving forward and inverse problems involving nonlinear partial differential equations.
\newblock \emph{Journal of Computational physics}, 378:\penalty0 686--707, 2019.

\bibitem[Ruthotto et~al.(2020)Ruthotto, Osher, Li, Nurbekyan, and Fung]{Liwu2020machine}
Lars Ruthotto, Stanley~J Osher, Wuchen Li, Levon Nurbekyan, and Samy~Wu Fung.
\newblock A machine learning framework for solving high-dimensional mean field game and mean field control problems.
\newblock \emph{Proceedings of the National Academy of Sciences}, 117\penalty0 (17):\penalty0 9183--9193, 2020.

\bibitem[Saelens et~al.(2019)Saelens, Cannoodt, Todorov, and Saeys]{saelens2019comparison}
Wouter Saelens, Robrecht Cannoodt, Helena Todorov, and Yvan Saeys.
\newblock A comparison of single-cell trajectory inference methods.
\newblock \emph{Nature biotechnology}, 37\penalty0 (5):\penalty0 547--554, 2019.

\bibitem[Schiebinger et~al.(2019)Schiebinger, Shu, Tabaka, Cleary, Subramanian, Solomon, Gould, Liu, Lin, Berube, et~al.]{waddingot}
Geoffrey Schiebinger, Jian Shu, Marcin Tabaka, Brian Cleary, Vidya Subramanian, Aryeh Solomon, Joshua Gould, Siyan Liu, Stacie Lin, Peter Berube, et~al.
\newblock Optimal-transport analysis of single-cell gene expression identifies developmental trajectories in reprogramming.
\newblock \emph{Cell}, 176\penalty0 (4):\penalty0 928--943, 2019.

\bibitem[Sha et~al.(2024)Sha, Qiu, Zhou, and Nie]{Tigon}
Yutong Sha, Yuchi Qiu, Peijie Zhou, and Qing Nie.
\newblock Reconstructing growth and dynamic trajectories from single-cell transcriptomics data.
\newblock \emph{Nature Machine Intelligence}, 6\penalty0 (1):\penalty0 25--39, 2024.

\bibitem[Shi et~al.(2022)Shi, Aihara, Li, and Chen]{landscape_Shi}
Jifan Shi, Kazuyuki Aihara, Tiejun Li, and Luonan Chen.
\newblock Energy landscape decomposition for cell differentiation with proliferation effect.
\newblock \emph{National Science Review}, 9\penalty0 (8):\penalty0 nwac116, 2022.

\bibitem[Shi et~al.(2024)Shi, De~Bortoli, Campbell, and Doucet]{shi2024diffusion}
Yuyang Shi, Valentin De~Bortoli, Andrew Campbell, and Arnaud Doucet.
\newblock Diffusion schr{\"o}dinger bridge matching.
\newblock \emph{Advances in Neural Information Processing Systems}, 36, 2024.

\bibitem[Sohl-Dickstein et~al.(2015)Sohl-Dickstein, Weiss, Maheswaranathan, and Ganguli]{sohl2015deep}
Jascha Sohl-Dickstein, Eric Weiss, Niru Maheswaranathan, and Surya Ganguli.
\newblock Deep unsupervised learning using nonequilibrium thermodynamics.
\newblock In \emph{International conference on machine learning}, pp.\  2256--2265. PMLR, 2015.

\bibitem[Song et~al.(2021)Song, Sohl-Dickstein, Kingma, Kumar, Ermon, and Poole]{song2020score}
Yang Song, Jascha Sohl-Dickstein, Diederik~P Kingma, Abhishek Kumar, Stefano Ermon, and Ben Poole.
\newblock Score-based generative modeling through stochastic differential equations.
\newblock In \emph{International Conference on Learning Representations}, 2021.

\bibitem[Tong et~al.(2020)Tong, Huang, Wolf, Van~Dijk, and Krishnaswamy]{trajectorynet}
Alexander Tong, Jessie Huang, Guy Wolf, David Van~Dijk, and Smita Krishnaswamy.
\newblock Trajectorynet: A dynamic optimal transport network for modeling cellular dynamics.
\newblock In \emph{International conference on machine learning}, pp.\  9526--9536. PMLR, 2020.

\bibitem[Tong et~al.(2023)Tong, Kuchroo, Gupta, Venkat, San~Juan, Rangel, Zhu, Lock, Chaffer, and Krishnaswamy]{tong2023unblanced}
Alexander Tong, Manik Kuchroo, Shabarni Gupta, Aarthi Venkat, Beatriz~P San~Juan, Laura Rangel, Brandon Zhu, John~G Lock, Christine~L Chaffer, and Smita Krishnaswamy.
\newblock Learning transcriptional and regulatory dynamics driving cancer cell plasticity using neural ode-based optimal transport.
\newblock \emph{bioRxiv}, pp.\  2023--03, 2023.

\bibitem[Tong et~al.(2024{\natexlab{a}})Tong, FATRAS, Malkin, Huguet, Zhang, Rector-Brooks, Wolf, and Bengio]{cfm_tong}
Alexander Tong, Kilian FATRAS, Nikolay Malkin, Guillaume Huguet, Yanlei Zhang, Jarrid Rector-Brooks, Guy Wolf, and Yoshua Bengio.
\newblock Improving and generalizing flow-based generative models with minibatch optimal transport.
\newblock \emph{Transactions on Machine Learning Research}, 2024{\natexlab{a}}.
\newblock ISSN 2835-8856.
\newblock Expert Certification.

\bibitem[Tong et~al.(2024{\natexlab{b}})Tong, Malkin, Fatras, Atanackovic, Zhang, Huguet, Wolf, and Bengio]{sflowmatch}
Alexander Tong, Nikolay Malkin, Kilian Fatras, Lazar Atanackovic, Yanlei Zhang, Guillaume Huguet, Guy Wolf, and Yoshua Bengio.
\newblock Simulation-free schr{\"o}dinger bridges via score and flow matching.
\newblock In \emph{International Conference on Artificial Intelligence and Statistics}, pp.\  1279--1287. PMLR, 2024{\natexlab{b}}.

\bibitem[Ventre et~al.(2023)Ventre, Forrow, Gadhiwala, Chakraborty, Angel, and Schiebinger]{Geofftrajectory}
Elias Ventre, Aden Forrow, Nitya Gadhiwala, Parijat Chakraborty, Omer Angel, and Geoffrey Schiebinger.
\newblock Trajectory inference for a branching sde model of cell differentiation.
\newblock \emph{arXiv preprint arXiv:2307.07687}, 2023.

\bibitem[Wan et~al.(2023)Wan, Zhang, Bao, Dong, and Shi]{DOngbin2023scalable}
Wei Wan, Yuejin Zhang, Chenglong Bao, Bin Dong, and Zuoqiang Shi.
\newblock A scalable deep learning approach for solving high-dimensional dynamic optimal transport.
\newblock \emph{SIAM Journal on Scientific Computing}, 45\penalty0 (4):\penalty0 B544--B563, 2023.

\bibitem[Wang et~al.(2021)Wang, Jiao, Xu, Wang, and Yang]{wang2021deep}
Gefei Wang, Yuling Jiao, Qian Xu, Yang Wang, and Can Yang.
\newblock Deep generative learning via schr{\"o}dinger bridge.
\newblock In \emph{International conference on machine learning}, pp.\  10794--10804. PMLR, 2021.

\bibitem[Wang et~al.(2010)Wang, Li, and Wang]{Lichunhe2010potential}
Jin Wang, Chunhe Li, and Erkang Wang.
\newblock Potential and flux landscapes quantify the stability and robustness of budding yeast cell cycle network.
\newblock \emph{Proceedings of the National Academy of Sciences}, 107\penalty0 (18):\penalty0 8195--8200, 2010.

\bibitem[Weinreb et~al.(2020)Weinreb, Rodriguez-Fraticelli, Camargo, and Klein]{weinreb2020lineage}
Caleb Weinreb, Alejo Rodriguez-Fraticelli, Fernando~D Camargo, and Allon~M Klein.
\newblock Lineage tracing on transcriptional landscapes links state to fate during differentiation.
\newblock \emph{Science}, 367\penalty0 (6479):\penalty0 eaaw3381, 2020.

\bibitem[Wu et~al.(2023)Wu, Liu, Ye, and Zhou]{zhouhao2024parameterized_H}
Hao Wu, Shu Liu, Xiaojing Ye, and Haomin Zhou.
\newblock Parameterized wasserstein hamiltonian flow.
\newblock \emph{arXiv preprint arXiv:2306.00191}, 2023.

\bibitem[Zhang et~al.(2024{\natexlab{a}})Zhang, Larschan, Bigness, and Singh]{scnode}
Jiaqi Zhang, Erica Larschan, Jeremy Bigness, and Ritambhara Singh.
\newblock {scNODE: generative model for temporal single cell transcriptomic data prediction}.
\newblock \emph{Bioinformatics}, 40\penalty0 (Supplement\_2):\penalty0 ii146--ii154, 09 2024{\natexlab{a}}.
\newblock ISSN 1367-4811.

\bibitem[Zhang et~al.(2024{\natexlab{b}})Zhang, Gao, Guo, and Duan]{Duan_action}
Peng Zhang, Ting Gao, Jin Guo, and Jinqiao Duan.
\newblock Action functional as early warning indicator in the space of probability measures.
\newblock \emph{arXiv preprint arXiv:2403.10405}, 2024{\natexlab{b}}.

\bibitem[Zhang et~al.(2021)Zhang, Afanassiev, Greenstreet, Matsumoto, and Schiebinger]{Stephenzhang2021optimal}
Stephen Zhang, Anton Afanassiev, Laura Greenstreet, Tetsuya Matsumoto, and Geoffrey Schiebinger.
\newblock Optimal transport analysis reveals trajectories in steady-state systems.
\newblock \emph{PLoS computational biology}, 17\penalty0 (12):\penalty0 e1009466, 2021.

\bibitem[Zhao et~al.(2024)Zhao, Zhang, and Li]{epr}
Yue Zhao, Wei Zhang, and Tiejun Li.
\newblock {EPR-Net: constructing a non-equilibrium potential landscape via a variational force projection formulation}.
\newblock \emph{National Science Review}, 11\penalty0 (7), 2024.

\bibitem[Zhou et~al.(2024{\natexlab{a}})Zhou, Lou, Khanna, and Ermon]{zhou2024denoising}
Linqi Zhou, Aaron Lou, Samar Khanna, and Stefano Ermon.
\newblock Denoising diffusion bridge models.
\newblock In \emph{The Twelfth International Conference on Learning Representations}, 2024{\natexlab{a}}.

\bibitem[Zhou et~al.(2024{\natexlab{b}})Zhou, Osher, and Li]{LIWUCHEN_score}
Mo~Zhou, Stanley Osher, and Wuchen Li.
\newblock Score-based neural ordinary differential equations for computing mean field control problems.
\newblock \emph{arXiv preprint arXiv:2409.16471}, 2024{\natexlab{b}}.

\bibitem[Zhou \& Li(2016)Zhou and Li]{landscape_PZ}
Peijie Zhou and Tiejun Li.
\newblock Construction of the landscape for multi-stable systems: Potential landscape, quasi-potential, a-type integral and beyond.
\newblock \emph{The Journal of chemical physics}, 144\penalty0 (9), 2016.

\bibitem[Zhou et~al.(2021{\natexlab{a}})Zhou, Gao, Li, Li, Niu, Ouyang, Lou, Li, and Li]{zhou2021stochasticity}
Peijie Zhou, Xin Gao, Xiaoli Li, Linxi Li, Caoyuan Niu, Qi~Ouyang, Huiqiang Lou, Tiejun Li, and Fangting Li.
\newblock Stochasticity triggers activation of the s-phase checkpoint pathway in budding yeast.
\newblock \emph{Physical Review X}, 11\penalty0 (1):\penalty0 011004, 2021{\natexlab{a}}.

\bibitem[Zhou et~al.(2021{\natexlab{b}})Zhou, Wang, Li, and Nie]{zhou2021dissecting}
Peijie Zhou, Shuxiong Wang, Tiejun Li, and Qing Nie.
\newblock Dissecting transition cells from single-cell transcriptome data through multiscale stochastic dynamics.
\newblock \emph{Nature communications}, 12\penalty0 (1):\penalty0 5609, 2021{\natexlab{b}}.

\bibitem[Zhou et~al.(2024{\natexlab{c}})Zhou, Bocci, Li, and Nie]{zhou2024spatial}
Peijie Zhou, Federico Bocci, Tiejun Li, and Qing Nie.
\newblock Spatial transition tensor of single cells.
\newblock \emph{Nature Methods}, pp.\  1--10, 2024{\natexlab{c}}.

\bibitem[Zhou et~al.(2024{\natexlab{d}})Zhou, Yu, and Li]{zhou2024revealing}
Ruisong Zhou, Yuguo Yu, and Chunhe Li.
\newblock Revealing neural dynamical structure of c. elegans with deep learning.
\newblock \emph{Iscience}, 27\penalty0 (5), 2024{\natexlab{d}}.

\end{thebibliography}
\bibliographystyle{iclr2025_conference}

\appendix
\section{Training Regularized Unbalanced Optimal Transport}
\subsection{\revisenew{Training} of Regularized Unbalanced Optimal Transport}\label{appen:learn_alg}
\revise{After the pre-training stage, once we have obtained initial $\boldsymbol{v}_\theta, g_\theta$ and $s_\theta$, then we can continue training by minimizing the total loss:
$$
\mathcal{L}= \mathcal{L}_{\text{Energy}}+\lambda_r\mathcal{L}_{\text{Recons}}+\lambda_f\mathcal{L}_{\text{FP}}.
$$
To compute the each component of the loss, 
the temporal integral and ODEs were numerically solved using the Neural ODE solver \citep{chen2018neuralode}. The gradients of the loss function with respect to the parameters in the neural networks for $\vxt$, $g(\boldsymbol{x},t)$ and $s(\boldsymbol{x},t)$ were computed using neural ODEs with a memory-efficient implementation. For computing the Wasserstein distance between discrete distributions, we utilize the implementation provided by the Python Optimal Transport library (POT) \citep{flamary2021pot}. \revisenew{The algorithm is detailed in \cref{algo:uddsb} (Training Stage).}
}

\subsection{Training Log Density Function}\label{appen:train_alg}
To utilize conditional flow matching (CFM) to learn an initial log density, first we choose pair samples $(\boldsymbol{x}_0, \boldsymbol{x}_1)$ from the optimal transport  plan $q(\boldsymbol{x}_0, \boldsymbol{x}_1)$ and then construct Brownian bridges between these pair samples. We first consider $\sigma(t)=\sigma$ is constant. The log density function is then matched with these Brownian bridges, i.e., $p(\boldsymbol{x}, t \mid (\boldsymbol{x}_0, \boldsymbol{x}_1) )=\mathcal{N}\left(\boldsymbol{x} ; t \boldsymbol{x}_1+(1-t) \boldsymbol{x}_0, \sigma^2 t(1-t)\right)$ and $\nabla_{\boldsymbol{x}} \log p(\boldsymbol{x}, t \mid (\boldsymbol{x}_0, \boldsymbol{x}_1) )=\frac{t \boldsymbol{x}_1+(1-t)\boldsymbol{x}_0-\boldsymbol{x}}{\sigma^2 t(1-t)}, ~ t \in [0, 1].$  We then utilize the fully connected neural networks $s_{\theta}(\boldsymbol{x},t)$ to approximate the log density function  $\frac{1}{2} \sigma^2 \log \pxt$ with a weighting function $\lambda_s$. Then we have 
$$
\mathcal{L}_{us}=\lambda_s^2\|\nabla_{\boldsymbol{x}} s_\theta-\frac{1}{2}\sigma^2 \nabla_{\boldsymbol{x}}\log \pxt\|_2^2.
$$
And the correspongding CFM loss is 
$$
\mathcal{L}_{\text{score}}=\mathbb{E}_{Q'}\lambda_s^2\|\nabla_{\boldsymbol{x}} s_\theta-\frac{1}{2}\sigma^2 \nabla_{\boldsymbol{x}}\log p (\boldsymbol{x},t \mid (\boldsymbol{x}_0, \boldsymbol{x}_1))\|_2^2,
$$
where $Q'=(t \sim \mathcal{U}(0,1)) \otimes q(\boldsymbol{x}_0, \boldsymbol{x}_1) \otimes p(\boldsymbol{x}, t \mid (\boldsymbol{x}_0, \boldsymbol{x}_1)) $.
We take the weighting function as
$$
\lambda_s(t)=\frac{2}{\sigma^2} \sigma \sqrt{t (1-t)}=\frac{2 \sqrt{t(1-t)}}{\sigma}.
$$

Then we have
\begin{equation}\label{eq:score}
    \begin{aligned}
\mathcal{L}_{\text{score}}=&\lambda_s(t)^2\left\|\nabla_{\boldsymbol{x}} s_{\theta}(\boldsymbol{x},t)-\frac{\sigma^2}{2}\nabla_{\boldsymbol{x}} \log p\left(\boldsymbol{x}\mid \boldsymbol{x_0}, \boldsymbol{x_1}\right)\right\|_2^2, \\& =\left\|\lambda_s(t) \nabla_{\boldsymbol{x}} s_{\theta}(\boldsymbol{x},t)-\lambda_s(t)\frac{\sigma^2}{2} \nabla_{\boldsymbol{x}} \log p\left(\boldsymbol{x}\mid \boldsymbol{x_0}, \boldsymbol{x_1}\right)\right\|_2^2, \\
& =\left\|\lambda_s(t) \nabla_{\boldsymbol{x}} s_{\theta}(\boldsymbol{x},t)+\boldsymbol{\epsilon}_1\right\|_2^2,
\end{aligned}
\end{equation}

where $\boldsymbol{\epsilon}_1 \sim \mathcal{N}(0,\boldsymbol{I})$.
This is also numerically stable. For the case where $\sigma(t)$ is not constant, a similar approach can be applied. One may refer to \citep{sflowmatch} for further details.

\subsection{Estimating Initial Distribution for Fokker-Planck Equation}\label{appen:pinn}
The initial distribution is estimated by 
a Gaussian mixture model to generate density 
$$
\begin{aligned}
p(\boldsymbol{x},0)=&p_{0}(\boldsymbol{x})=\sum_{i=1}^{N_0} \frac{\exp \left(-\frac{1}{2}\left(\boldsymbol{x}-\boldsymbol{x}_i (t_0)\right)^T \boldsymbol{\Sigma}^{-1}\left(\boldsymbol{x}-\boldsymbol{x}_i (t_0)\right)\right)}{\sqrt{(2 \pi)^d|\boldsymbol{\Sigma}|}}, \boldsymbol{\Sigma}=\sigma \boldsymbol{I} \in \mathbb{R}^{d \times d}.
\end{aligned}
$$
\subsection{Energy Loss}\label{appendix:energy}
\revise{In \citep{Tigon}, it introduces the following theorem to transform the high-dimensional integral to the Monte Carlo integral.}
\begin{theorem}\label{thm:energy}
\revise{
    If smooth density $p(\boldsymbol{x},t): \mathbb{R}^d \times[0,1] \rightarrow \mathbb{R}^{+}$, velocity field $\vxt: \mathbb{R}^d \times[0,1] \rightarrow \mathbb{R}^d$ and growth rate $g(\boldsymbol{x}, t): \mathbb{R}^d \times[0,1] \rightarrow \mathbb{R}$ satisfy
$$
\left\{\begin{array}{c}
\partial_t p(\boldsymbol{x},t)+\nabla \cdot(\vxt p(\boldsymbol{x},t))=\gxt p(\boldsymbol{x},t), \\
p(\boldsymbol{x}, 0)=p_0(\boldsymbol{x}),
\end{array}\right.
$$
for all $0 \leq t \leq 1$ with $\frac{\mathrm{d} \boldsymbol{x}(t)}{\mathrm{d} t}=\vxt$ and  $\boldsymbol{x}(0)=\boldsymbol{x}_0$,  then for any measurable function $f(\boldsymbol{x}, t): \mathbb{R}^d \times[0,1] \rightarrow \mathbb{R}^d$, we have
$
\int_0^1 \int_{\mathbb{R}^d} f(\boldsymbol{x}, t) p(\boldsymbol{x},t) \mathrm{d}\boldsymbol{x} \mathrm{d} t=\mathbb{E}_{\boldsymbol{x_0} \sim p_0} \int_0^1 f(\boldsymbol{x}, t) e^{\int_0^tg(\boldsymbol{x}, s) \rmd s} \mathrm{d} t .
$
}
\end{theorem}

\section{Additional Results}
\subsection{Synthetic Gene Regulatory Network}\label{appen:gene}
The system dynamics are governed by the following set of stochastic ordinary differential equations (ODEs):
$$
\begin{aligned}
& \frac{\mathrm{d} X_1}{\mathrm{~d} t} = \frac{\alpha_1 X_1^2 + \beta}{1 + \alpha_1 X_1^2 + \gamma_2 X_2^2 + \gamma_3 X_3^2 + \beta} - \delta_1 X_1 + \eta_1 \xi_t, \\
& \frac{\mathrm{d} X_2}{\mathrm{~d} t} = \frac{\alpha_2 X_2^2 + \beta}{1 + \gamma_1 X_1^2 + \alpha_2 X_2^2 + \gamma_3 X_3^2 + \beta} - \delta_2 X_2 + \eta_2 \xi_t, \\
& \frac{\mathrm{d} X_3}{\mathrm{~d} t} = \frac{\alpha_3 X_3^2}{1 + \alpha_3 X_3^2} - \delta_3 X_3 + \eta_3 \xi_t.
\end{aligned}
$$

Genes $X_1$ and $X_2$ mutually inhibit each other and self-activate, forming a toggle switch. An external signal $\beta$ uniformly activates both $X_1$ and $X_2$ independently of gene expression levels. Gene $X_3$ inhibits the expression of both $X_1$ and $X_2$. Here, $X_i(t)$ represents the concentration of gene $i$ at time $t$, with $\alpha_i$ and $\gamma_i$ denoting the strengths of self-activation and inhibition, respectively. The parameters $\delta_i$ represent gene degradation rates, while $\eta_i \xi_t$ accounts for the stochastic effects with additive white noise. The probability of cell division correlates positively with the expression of $X_2$, calculated as $g =\alpha_g \frac{X_2^2}{1 + X_2^2} \%$. Upon division,  cells inherit the gene expression states $(X_1(t), X_2(t), X_3(t))$ of the parent cell, subject to independent perturbations $\eta_d \mathcal{N}(0,1)$ for each gene, and transition independently thereafter. The hyper-parameters are listed at \cref{tab:simulation_parameters}. The initial cells are chosen independently and identically distributed from two normal distributions $\mathcal{N}([2,0.2,0],0.01)$ and $\mathcal{N}([0,0,2],0.01)$. At each step, we corrected the negative expression value to 0. 

\begin{table}[ht]
    \centering
    \caption{Simulation parameters on gene regulatory network.}
    \label{tab:simulation_parameters}
    \begin{tabular}{lll}
        \hline
        \textbf{Parameter} & \textbf{Value} & \textbf{Description} \\
        \hline
        $\alpha_1$ & 0.5 & Strength of self-activation for $X_1$ \\
        $\gamma_1$ & 0.5 & Strength of inhibition by $X_3$ on $X_1$ \\
        $\alpha_2$ & 1 & Strength of self-activation for $X_2$ \\
        $\gamma_2$ & 1 & Strength of inhibition by $X_3$ on $X_2$ \\
        $\alpha_3$ & 1 & Strength of self-activation for $X_3$ \\
        $\gamma_3$ & 10 & Half-saturation constant for inhibition terms \\
        $\delta_1$ & 0.4 & Degradation rate for $X_1$ \\
        $\delta_2$ & 0.4 & Degradation rate for $X_2$ \\
        $\delta_3$ & 0.4 & Degradation rate for $X_3$ \\
        $\eta_1$ & 0.05 & Noise intensity for $X_1$ \\
        $\eta_2$ & 0.05 & Noise intensity for $X_2$ \\
        $\eta_3$ & 0.01 & Noise intensity for $X_3$ \\
        $\eta_d$ & 0.014 & Noise intensity for cell perturbations \\
        $\beta$ & 1 & External signal activating $X_1$ and $X_2$ \\
        $dt$ & 1 & Time step size \\
        Time Points & [0, 8, 16, 24, 32] & Time points at which data is recorded \\
        \hline
    \end{tabular}
\end{table}

\begin{figure}[htp]
    \centering
    \includegraphics[width=1\linewidth]{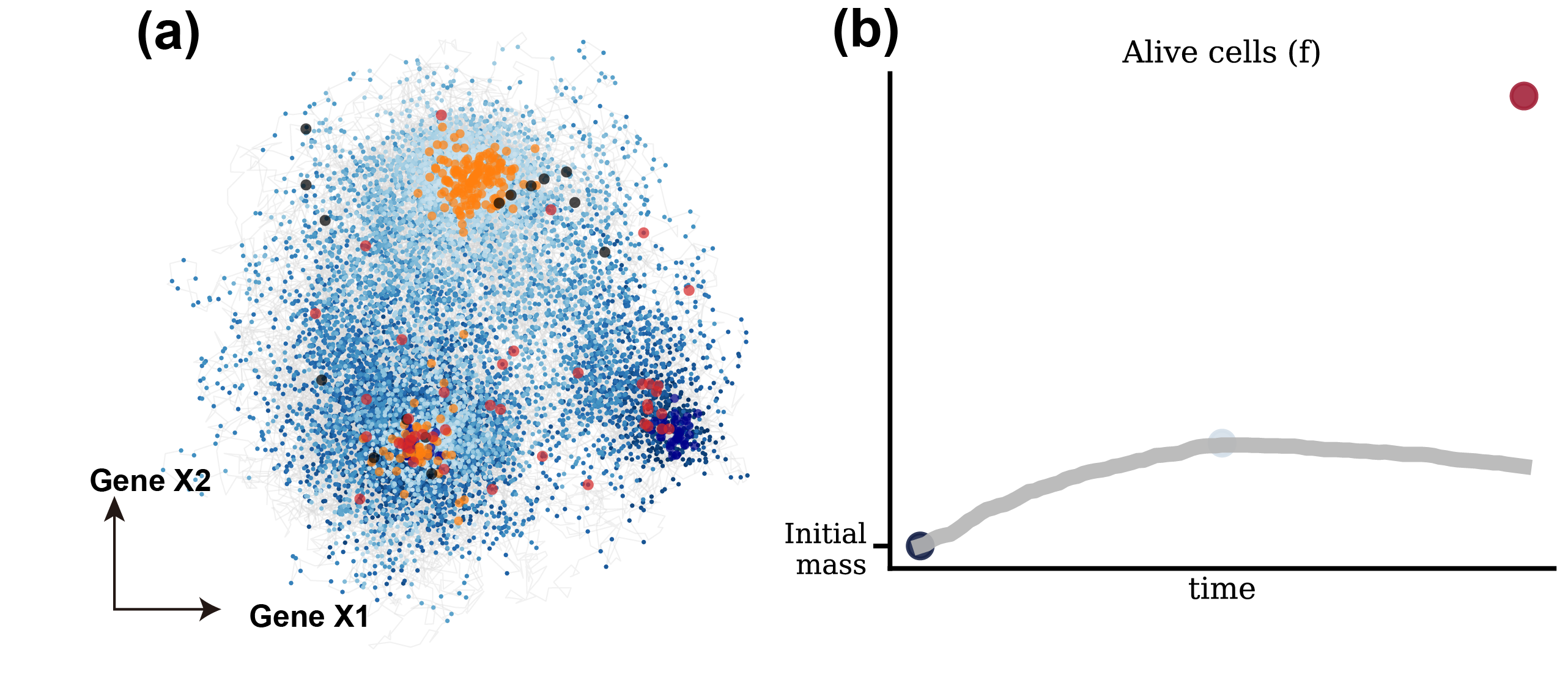}
    \caption{\revise{{\bf Results obtained by UDSB \citep{bunne_unsb} on gene regulatory network.} (a) The trajectory learned by UDSB, where black dots indicate particle death, red dots signify particle growth, orange dots represent the target distribution, dark blue dots denote the source distribution, and gradient blue dots illustrate particle trajectories.
(b) Predicted changes in cell population at intermediate time points, with dots representing the actual mass.}}
    \label{fig:bunnegene}
\end{figure}

\revise{We conducted a comprehensive evaluation of the methodologies proposed by \citep{action_matching,bunne_unsb} utilizing our simulated dataset. As detailed in \cref{table:gene}, our approach consistently outperforms the referenced methods across multiple quantitative metrics, thereby demonstrating its superior ability to capture the underlying dynamics of the system.  Furthermore, \cref{fig:bunnegene} provides  the outcomes produced by the UDSB \citep{bunne_unsb}. While their method  predicts an overall increase in cell population, a closer examination reveals the presence of false dynamics within the predicted transitions. }

\subsection{Synthetic Gaussian Mixtures}\label{appen:gauss}
For the initial distribution, we generated 400 samples from the Gaussian located lower in the $(x_1, x_2)$ plane, and 100 samples from the Gaussian positioned higher. For the final distribution, we generated 1,000 samples from the upper Gaussian, and 200 samples each from the two lower Gaussians. We then tested the RUOT-based model’s ability to learn the stochastic dynamics using the samples generated in this manner.
\begin{figure}[htp]
    \centering
    \includegraphics[width=0.9\linewidth]{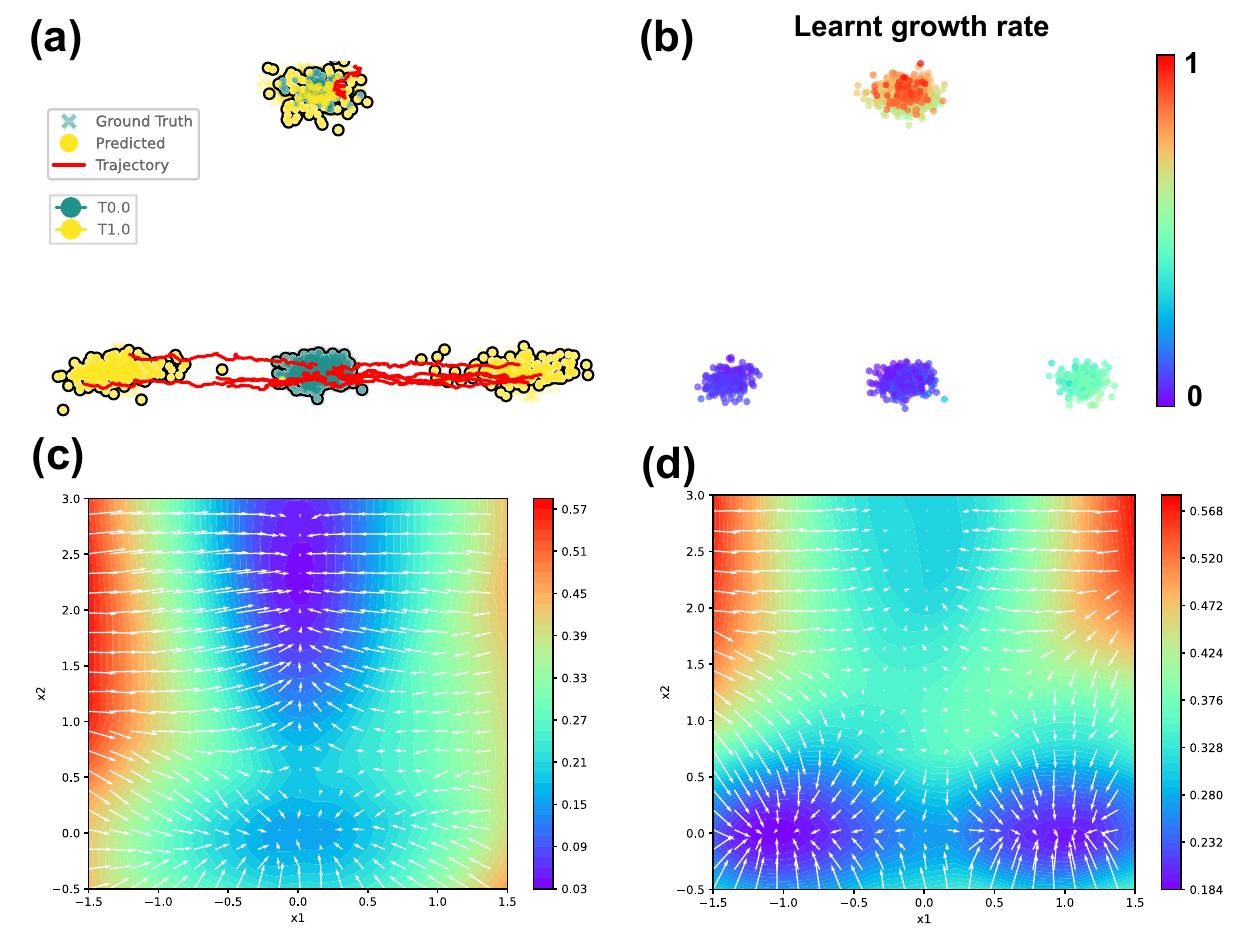}
    \caption{{\bf Results of DeepRUOT on Gaussian mixtures ($\sigma=0.1$, 10D).} (a) The learned trajectory by DeepRUOT ($\sigma=0.1$). (b) The growth rate inferred by our model. (c) The Waddington developmental landscape at $t=0$ ($\sigma=0.1$). (d) The Waddington developmental landscape at $t=1$ ($\sigma=0.1$). }
    \label{fig:gauss_10d}
\end{figure}

\begin{figure}[htp]
    \centering
    \includegraphics[width=1\linewidth]{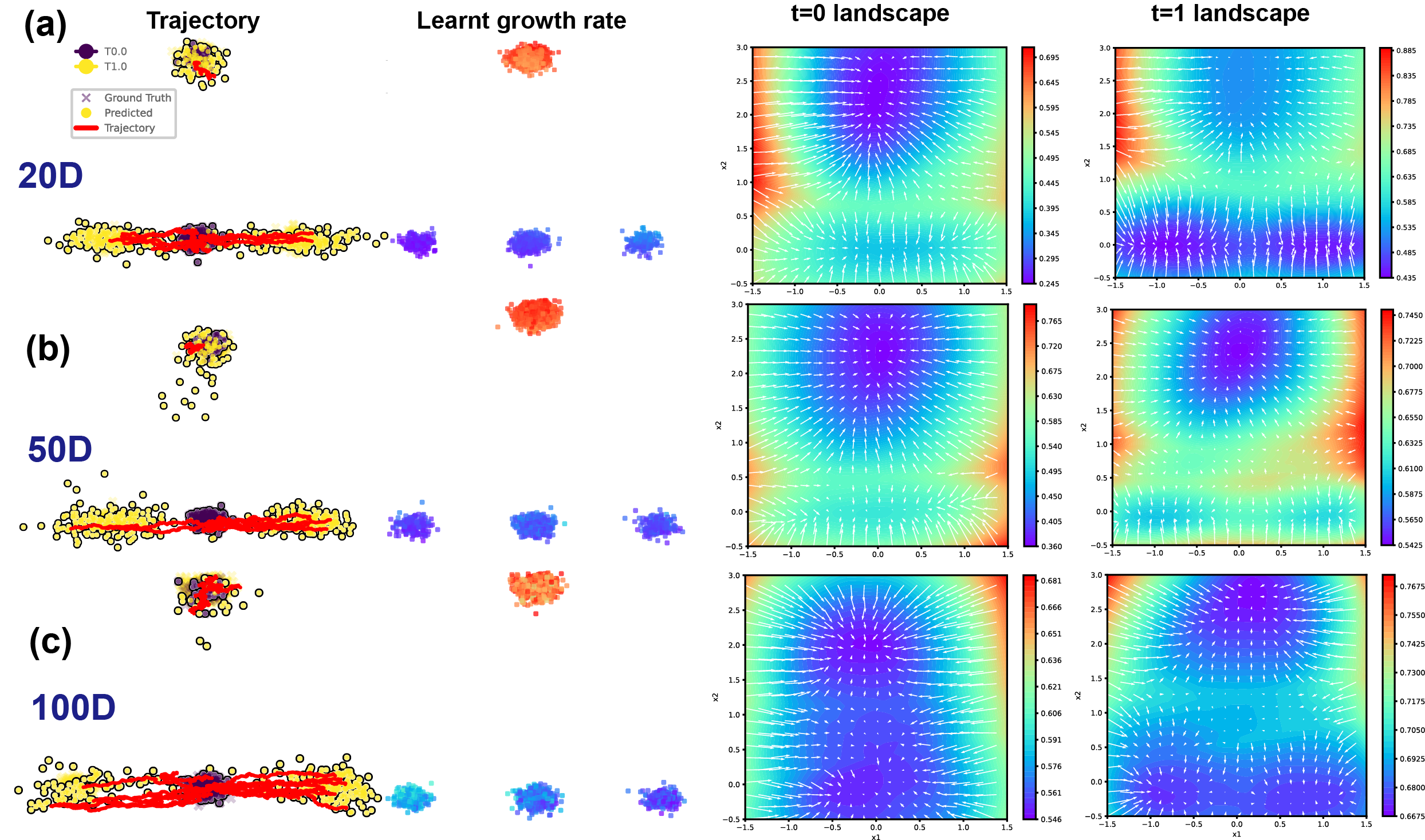}
    \caption{\revise{{\bf Results of DeepRUOT on higher dimensional Gaussian mixtures ($\sigma=0.1$).} The first column presents the trajectories learned by DeepRUOT ($\sigma = 0.1$), the second column displays the growth rates inferred by our model, the third column shows the Waddington developmental landscape at $t = 0$ ($\sigma = 0.1$), and the fourth column depicts the Waddington developmental landscape at $t = 1$ ($\sigma = 0.1$).
(a) 20 dimensions,
(b) 50 dimensions,
(c) 100 dimensions.}}
    \label{fig:highgauss}
\end{figure}

\revise{To further evaluate the scalability of our algorithm, we simulated Gaussian mixture models with 20, 50, and 100 dimensions, as illustrated in \cref{fig:highgauss}. We similarly visualize the results on the first and second dimensions. The results demonstrate that our model remains effective and applicable across these higher-dimensional settings. }

\revise{Similar to the frameworks such as TrajectoryNet \citep{trajectorynet}, MIOflow \citep{mioflow} and \citep{dyn_sb_koshizuka2023neural}, which involve simulating an ODE/SDE and performing optimal transport for distribution matching, the computational cost and scalability of our method are comparable to these existing approaches. Notably, the score matching in our method is conducted via conditional flow matching \citep{cfm_lipman,cfm_tong,sflowmatch}, which is simulation-free and thus highly efficient.  Consequently, our approach does not introduce significant additional computational overhead, since the cost of each component is similar to previous works.  Meanwhile, when it extends to thousands of dimensions, our approach similarly encounters challenges, necessitating the development of simulation-free training methods akin to flow matching to handle higher-dimensional settings effectively \citep{sflowmatch,cfm_lipman,cfm_tong}. However, in the field of single-cell biology, dimensionality reduction of gene expression data is routinely employed, making hundreds of dimensional space sufficient for practical applications.}

\subsection{Single Cell Dynamics in Mouse Hematopoiesis}\label{appen:sc-mouse}

\begin{figure}[htp]
    \centering
    \includegraphics[width=0.9\linewidth]{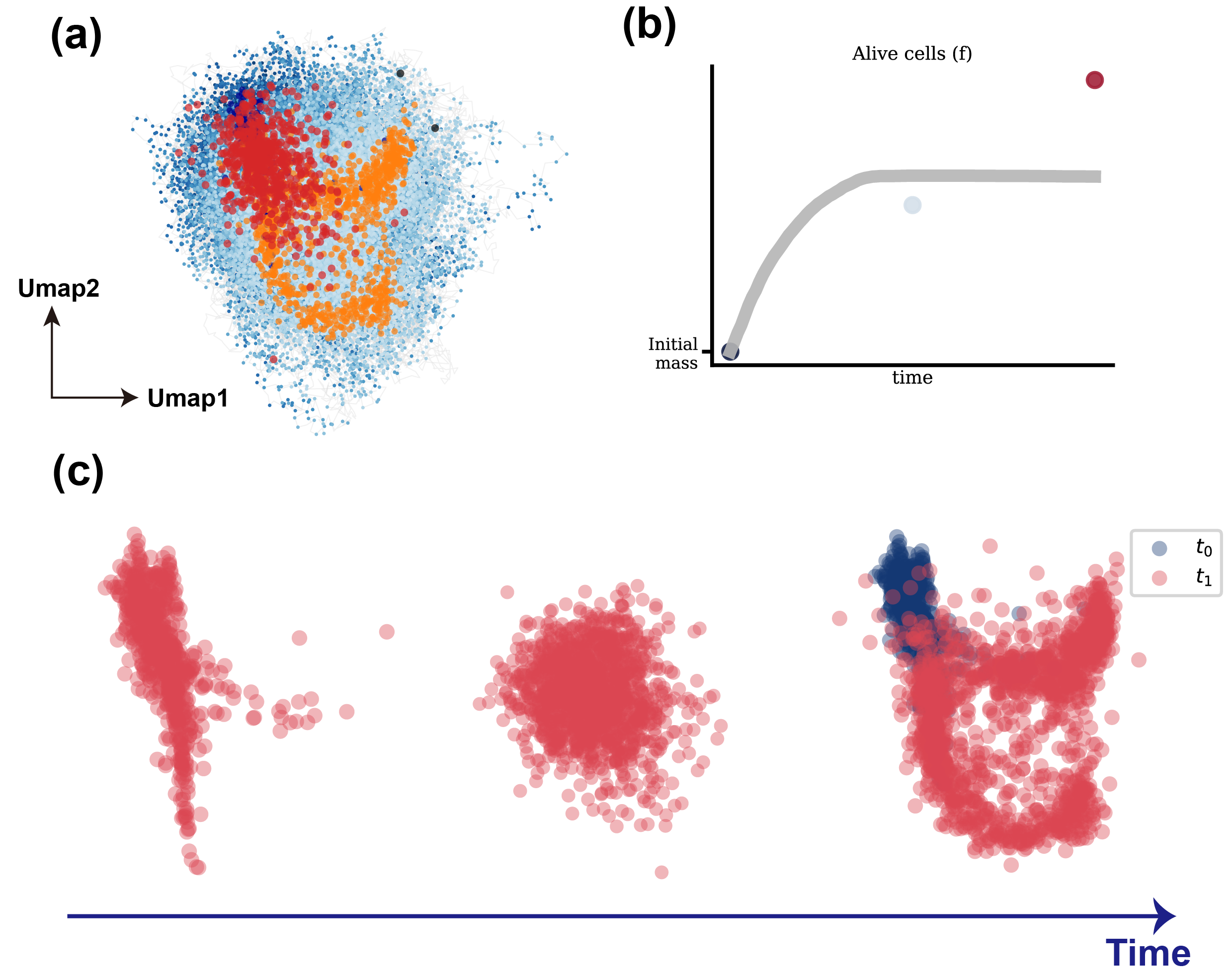}
    \caption{\revise{ {\bf Results obtained by UDSB \citep{bunne_unsb} on mouse hematopoiesis.} (a) The trajectory learned by UDSB, where black dots indicate particle death, red dots signify particle growth, orange dots represent the target distribution, dark blue dots denote the source distribution, and gradient blue dots illustrate particle trajectories.
(b) Predicted changes in cell population at intermediate time points, with dots representing the actual mass.  (c) Red denotes the predicted cell distributions at various time points, while the blue distribution at the final time point represents the initial distribution.}}
    \label{fig:bunnescRNA}
\end{figure}

\revise{We conducted a comparative evaluation of the methodologies proposed by \citep{action_matching,bunne_unsb} using our mouse hematopoiesis dataset. As detailed in \cref{table:gene}, our approach consistently outperforms the referenced methods across a range of quantitative metrics, underscoring its superior efficacy in capturing the underlying biological dynamics. In \cref{fig:bunnescRNA}, we present the results generated by the UDSB \citep{bunne_unsb}. \cref{fig:bunnescRNA}(a) displays the learned trajectory, where black dots indicate particle death, red dots signify particle growth, orange dots represent the target distribution, dark blue dots denote the source distribution, and gradient blue lines illustrate the particle trajectories. This visualization highlights the dynamic trajectory of cell population changes over time. \cref{fig:bunnescRNA}(b) illustrates the predicted changes in cell population at intermediate time points, with each dot representing the actual mass observed.  Finally, \cref{fig:bunnescRNA}(c) shows the predicted cell distributions at various time points in red, while the blue distribution at the final time point represents the initial distribution. This panel underscores the model’s ability to track and predict the evolution of cell distributions from the initial state to subsequent stages.
}

\revise{
Notably, although this method achieves a marginally higher $\mathcal{W}_1$ metric at the final time point, a closer examination reveals that the predicted distributions at intermediate time points, as illustrated in \cref{fig:bunnescRNA}(c), deviate  from the true data distributions. This indicates that the \citep{bunne_unsb} method may have challenges to accurately model the transitional dynamics that occur between the initial and final stages of hematopoiesis.}

\revise{Furthermore,  \cref{fig:bunnescRNA}(a) and \cref{fig:bunnescRNA}(b) demonstrate that the growth locations predicted by the UDSB algorithm are consistent with those identified by our model. This consistency provides a form of cross-validation, affirming that our algorithm effectively captures both the growth and migration dynamics inherent in the cell population. So the ability of our model to maintain accurate predictions across all time points, including the  intermediate stages, demonstrates its enhanced capacity for modeling unbalanced dynamics within complex biological systems. }

\subsection{Single Cell Dynamics in EMT}\label{appen:scRNA-emt}
\revise{Subsequently, we applied DeepRUOT to a time-series single-cell RNA sequencing  dataset derived from the A549 cancer cell line. In this study, cells were treated with TGFB1 to induce epithelial-mesenchymal transition (EMT) during the initial four-time points \citep{Tigon}. Cells harvested at each time point were cultured in vitro with identical initial cell numbers, ensuring that the cell counts directly reflect the dynamics of the cell population over time.
}

\revise{
We employed Principal Component Analysis (PCA) to reduce the dimensionality of our single-cell RNA sequencing (scRNA-seq) data to a ten-dimensional latent space. This latent representation was subsequently utilized as the input for DeepRUOT. To aid in the interpretation of the results, we projected the algorithm’s outputs onto the first and second principal components, which capture the majority of the variance in the data and thus provide meaningful insights into the underlying biological phenomena.
}

\revise{
The trajectories inferred by DeepRUOT exhibited consistent and biologically plausible transition dynamics, effectively mapping the progression of cells through different developmental states. Furthermore, we visualized the growth rates estimated by the algorithm alongside the developmental landscapes at various time points, providing a comprehensive view of both the quantitative and qualitative aspects of cellular dynamics. 
}

\begin{figure}[htp]
    \centering
    \includegraphics[width=1\linewidth]{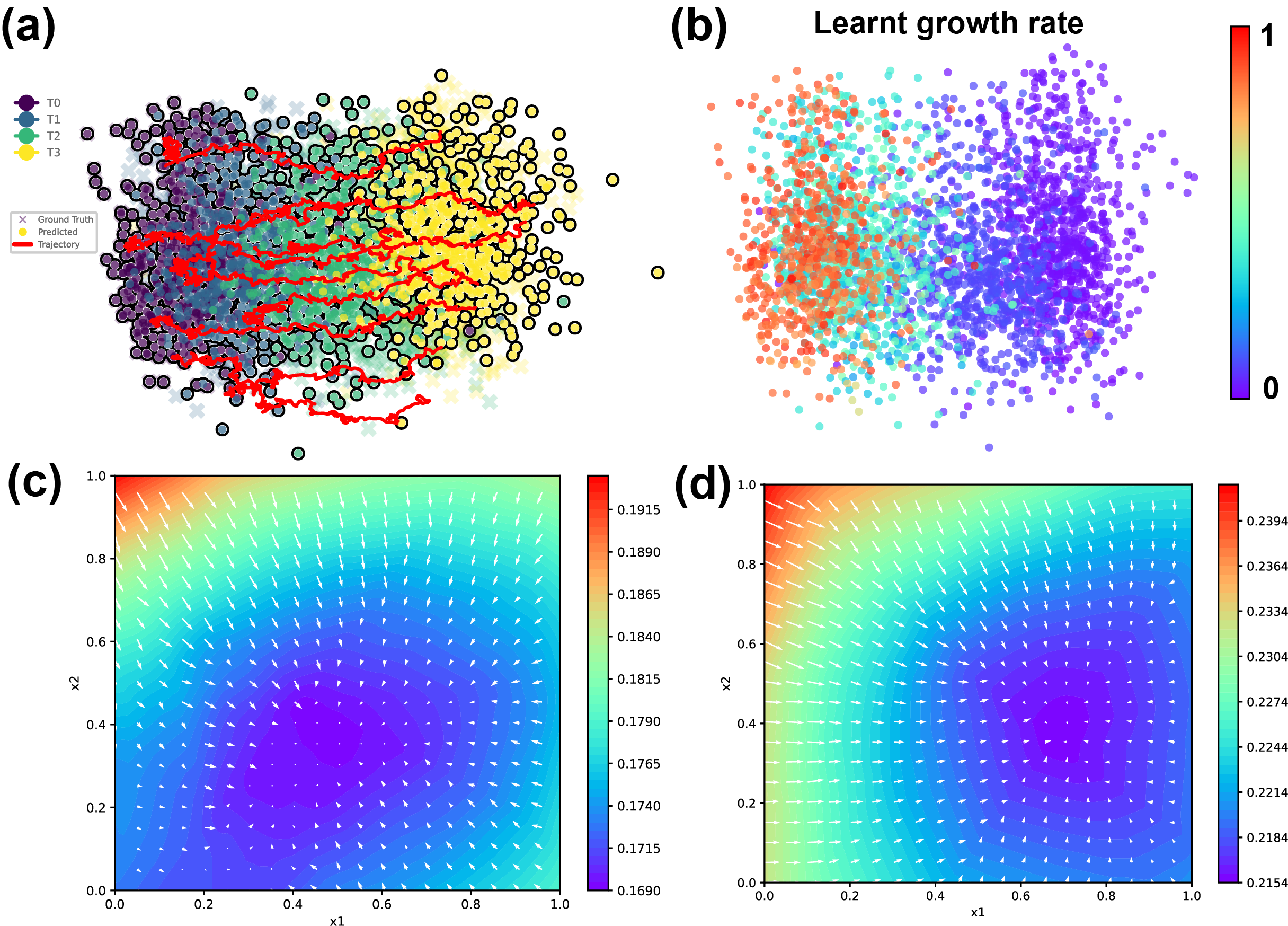}
    \caption{\revise{ {\bf Results of DeepRUOT on EMT scRNA-seq data ($\sigma=0.05$).} (a) The stochastic dynamics learned by RUOT ($\sigma=0.05$). (b) The growth rates learned by DeepRUOT ($\sigma=0.05$). (c) The constructed Waddington developmental landscape at $t=1$ ($\sigma=0.05$). (d) The landscape at $t=2$ ($\sigma=0.05$).}}
    \label{fig:emt}
\end{figure}

\revise{
Notably, the growth patterns inferred by DeepRUOT displayed elevated values during the initial and intermediate stage of epithelial-mesenchymal transition (EMT) compared to the epithelial (E) and mesenchymal (M) stages, as illustrated in \cref{fig:emt}. This observation aligns with previous studies that have reported enhanced stemness and proliferative capacity in cells at the intermediate stage.}

\revise{
Moreover, our landscape analysis revealed that the inferred developmental landscapes are consistent with the observed data distributions at corresponding time points. The developmental landscapes provide a representation of the cell states, illustrating how cells traverse through different regions of the latent space as they progress through various stages of differentiation and migration.}
\subsection{Ablation Studies}\label{appen:ablation}
\revise{We conducted ablation experiments on the gene regulatory dataset to evaluate the contributions of key components in our algorithm. Specifically, we examined the impact of the growth term  $\gxt$. Then we assessed the necessity of the mass loss term  $\mathcal{L}_\text{Mass}$  within the reconstruction loss function  $\mathcal{L}_{\text{Recons}}$ and $\mathcal{L}_\text{FP}$ in the total loss \eqref{eq:total_loss}. These analyses demonstrate the critical roles that the growth term, the mass loss component, and the Fokker-Planck constraint play in enhancing the performance and accuracy of our model.
}

\begin{table}[th]
\centering
\color{black}
\caption{Ablation studies   across five runs on gene regulatory data ($\sigma=0.25$). We show the mean value with one standard deviation.}
\begin{threeparttable}
\resizebox{\textwidth}{!}{ % Resize table to fit the width of the text
\begin{tabular}{lcccccccc}
\toprule
& \multicolumn{2}{c}{  $t=1$} & \multicolumn{2}{c}{  $t=2$} & \multicolumn{2}{c}{  $t=3$} & \multicolumn{2}{c}{  $t=4$} \\
\cmidrule(r){2-3} \cmidrule(r){4-5} \cmidrule(r){6-7} \cmidrule(r){8-9}
Model & $\mathcal{W}_1$ & $\mathcal{W}_2$ & $\mathcal{W}_1$ & $\mathcal{W}_2$ & $\mathcal{W}_1$ & $\mathcal{W}_2$ & $\mathcal{W}_1$ & $\mathcal{W}_2$ \\
\midrule
SF2M \citep{sflowmatch} & $0.174_{\pm 0.010}$ & $0.303_{\pm 0.023}$ & $0.430_{\pm 0.027}$ & $0.719_{\pm 0.032}$ & $0.686_{\pm 0.054}$ & $1.050_{\pm 0.047}$ & $0.871_{\pm 0.072}$ & 
$1.242_{\pm 0.065}$ \\
\revise{DeepRUOT w/o growth}   & \revise{$0.173_{\pm 0.005}$} & \revise{$0.198_{\pm 0.004}$} & \revise{${0.324}_{\pm 0.007}$} & \revise{${0.371}_{\pm 0.005}$}& \revise{$0.481_{\pm 0.010}$} & 
\revise{$0.576_{\pm 0.004}$} & 
\revise{${0.772}_{\pm 0.009}$} &
\revise{${0.877}_{\pm 0.011}$}
\\
\revise{DeepRUOT w/o $\mathcal{L}_\text{Mass}$}   & \revise{$0.159_{\pm 0.006}$} & \revise{$0.186_{\pm 0.006}$} & \revise{$0.308_{\pm 0.005}$} & \revise{$0.349_{\pm 0.007}
$} & \revise{${0.465}_{\pm 0.011}$} & \revise{${0.553}_{\pm 0.011}$} & \revise{$0.690_{\pm 0.007}$} & \revise{${0.816}_{\pm 0.009}$} \\
DeepRUOT w/o $\mathcal{L}_\text{FP}$
&${0.090}_{\pm 0.004}$ & $\boldsymbol{0.107}_{\pm 0.005}$ & 
 ${0.108}_{\pm 0.003}$ & ${0.132}_{\pm 0.002}$ & ${0.133}_{\pm 0.006}$ & $\boldsymbol{0.156}_{\pm 0.007}$ & ${0.165}_{\pm 0.008}$ & ${0.197}_{\pm 0.011}$ \\
 \revisenew{DeepRUOT w/o pre-training}& \revisenew{${0.635}_{\pm 0.006}$} &  \revisenew{${0.737}_{\pm 0.009}$} & 
  \revisenew{${1.113}_{\pm 0.017}$} &  \revisenew{${1.310}_{\pm 0.017}$} &  \revisenew{${1.329}_{\pm 0.020}$} &  \revisenew{${1.545}_{\pm 0.025}$} &  \revisenew{${1.366}_{\pm 0.023}$} &  \revisenew{${1.565}_{\pm 0.028}$}
 \\ 
\revisenew{DeepRUOT w/o score matching}& \revisenew{${1.291}_{\pm 0.007}$} &  \revisenew{${1,426}_{\pm 0.009}$} & 
  \revisenew{${2.051}_{\pm 0.003}$} &  \revisenew{${2.187}_{\pm 0.004}$} &  \revisenew{${2.675}_{\pm 0.010}$} &  \revisenew{${2.798}_{\pm 0.009}$} &  \revisenew{${3.290}_{\pm 0.009}$} &  \revisenew{${3.391}_{\pm 0.007}$}
 \\ 
  \revisenew{DeepRUOT w/o training}& \revisenew{$\boldsymbol{0.089}_{\pm 0.004}$} &  \revisenew{${0.108}_{\pm 0.005}$} & 
  \revisenew{$\boldsymbol{0.102}_{\pm 0.002}$} &  \revisenew{${0.132}_{\pm 0.012}$} &  \revisenew{${0.134}_{\pm 0.007}$} &  \revisenew{${0.158}_{\pm 0.013}$} &  \revisenew{${0.163}_{\pm 0.009}$} &  \revisenew{${0.200}_{\pm 0.015}$}
 \\
DeepRUOT& ${0.095}_{\pm 0.005}$ & ${0.115}_{\pm 0.006}$ & ${0.104}_{\pm 0.002}$ & $\boldsymbol{0.130}_{\pm 0.007}$ & $\boldsymbol{0.130}_{\pm 0.003}$ & ${0.157}_{\pm 0.003}$ & $\boldsymbol{0.140}_{\pm 0.007}$ & $\boldsymbol{0.168}_{\pm 0.007}$ \\
\bottomrule
\end{tabular}
} % End of resizebox
  \end{threeparttable}
  \label{table:ablation}
\end{table}

\revise{
We observe that when our algorithm does not account for growth, i.e., by setting  $\gxt = 0$, it reduces to regularized optimal transport, which is equivalent to the Schrödinger bridge problem. Consequently, SF2M \citep{sflowmatch} corresponds to the scenario where $\gxt = 0$ within our approach. Our experiments have demonstrated that omitting the growth factor leads to inaccurate trajectory reconstructions, underscoring the necessity of modeling unbalanced dynamics. To further substantiate this finding, we conducted an explicit ablation study with $\gxt$  (\cref{table:ablation}). The results from this experiment align with those of the SF2M, confirming that disabling the growth rate adversely affects performance by neglecting essential growth and death processes. This comparison reinforces the importance of the growth rate component in accurately capturing unbalanced stochastic dynamics, thereby validating the enhanced performance of DeepRUOT.
}

\revise{Next, we assess the impact of the mass loss term, $\mathcal{L}_{\text{Mass}}$, within the reconstruction loss function $\mathcal{L}_{\text{Recons}} = \lambda_m \mathcal{L}_{\text{Mass}} + \lambda_d \mathcal{L}_{\text{OT}}$ during all training stage. By setting $\lambda_m = 0$, we observed that the exclusion of this component results in inaccurate trajectory reconstructions (see  \cref{table:ablation}). This finding underscores the crucial role of incorporating local mass matching in unbalanced settings, demonstrating that the omission of $\mathcal{L}_{\text{Mass}}$ significantly impairs the model’s ability to accurately capture the system’s dynamics.}

\revise{
Subsequently, we investigate the \revisenew{training} stage using the loss function
$\mathcal{L} =   \mathcal{L}_{\text{Energy}} + \lambda_r \mathcal{L}_{\text{Recons}} + \lambda_f \mathcal{L}_{\text{FP}}.$
By setting $\lambda_f = 0$, thereby excluding the Fokker-Planck constraint, the results presented in \cref{table:ablation} reveal that incorporating this loss term enhances our model’s performance, particularly at the final time point. These results collectively highlight the importance of both mass matching and the Fokker-Planck constraint in improving the accuracy and reliability of trajectory reconstructions within our framework.}

\revisenew{Furthermore, we investigated the impact of different stages in our training process on the final results. Our training algorithm (\cref{algo:uddsb}) comprises two stages: the pre-training stage and the training stage. To assess their respective contributions, we conducted experiments under three scenarios: (1) omitting the pre-training stage, (2) excluding score matching during the pre-training stage, and (3) removing the training stage. As presented in \cref{table:ablation}, we observed that the algorithm failed to converge in the first two scenarios. Specifically, scenario (2) yielded poorer performance due to inaccurate score matching, which caused particle trajectories to deviate from the correct dynamical paths. In scenario (1), the necessity to balance multiple loss functions resulted in the vector field and score not being adequately learned, leading particles are inclined to remain near their initial positions. In scenario (3), the absence of the training stage slightly degraded the performance metrics, especially at later time points. These ablation studies demonstrate that the pre-training stage is critical within our framework, while the training stage contributes to the refinement of final results, ensuring we are addressing the regularized optimal transport problem, which is theoretically necessary. Empirical results indicate that the pre-training stage significantly enhances the performance of our algorithm.}

\section{Experiential Details}
\subsection{Evaluation Metrics}\label{appen:evalu}
We evaluate the empirical 1-Wasserstein distance ($\mathcal{W}_1$) and 2-Wasserstein distance ($\mathcal{W}_2$) on the gene data and scRNA-seq data. It is defined as follows:
$$
\mathcal{W}_2(p, q)=\left(\min _{\pi \in \boldsymbol{\Pi}\left(p, q\right)} \int\|\boldsymbol{x}-\boldsymbol{y}\|_2^2 d \pi(\boldsymbol{x}, \boldsymbol{y})\right)^{1 / 2},
$$
and
$$
\mathcal{W}_1(p, q)=\left(\min _{\pi \in \boldsymbol{\Pi}\left(p, q\right)} \int\|\boldsymbol{x}-\boldsymbol{y}\|_2 d \pi(\boldsymbol{x}, \boldsymbol{y})\right),
$$
where $p$ and $q$ represent empirical distributions. 

We evaluate our method by first learning the dynamics using data from all time points. We then apply these inferred dynamics to the initial data point to generate data for subsequent time points. Next, we compute the $\mathcal{W}_1$ and $\mathcal{W}_2$ distances between the generated data and the real data. For the gene regulatory network data, we ensure that the number of generated data points matches that of the real data since we have the true dynamics. In the scRNA-seq dataset, since the true dynamics are unknown, we apply the inferred dynamics to the initial data and compute the $\mathcal{W}_1$ and $\mathcal{W}_2$ distances between the generated data and all the data at that time point. Because our method could assign weights to the sample points, we compute the weighted $\mathcal{W}_1$ and $\mathcal{W}_2$  distances accordingly in the evaluation of our methods. We take the weight computed by DeepRUOT  without diffusion to use for simplicity. For other methods we used the default uniform distribution to evaluate.  

To evaluate the performance of SF2M and DeepRUOT on scRNA-seq datasets we take the same diffusion coefficient as $\sigma=0.25$. To evaluate the performance on synthetic gene regulatory data, we choose $\sigma=0.05$ for our DeepRUOT method and $\sigma=0.25$ for SF2M. This is due to we observe some numerical instability when using SF2M with $\sigma=0.05$. So to ensure a fair comparison, we choose the parameter as default in SF2M.

\revise{To evaluate the performance of \citep{Tigon} on gene regulatory network data and single-cell RNA sequencing (scRNA-seq) mouse hematopoiesis data, as we observed certain instabilities on the datasets,  to ensure a fair comparison,  we independently reimplemented the method and conducted additional evaluations (by setting $\sigma=0$ in DeepRUOT).}

\revise{To evaluate the performance of \citep{action_matching} and \citep{bunne_unsb} on gene regulatory network data and single-cell RNA sequencing (scRNA-seq) mouse hematopoiesis data, we utilized their default parameter settings. Specifically, for \citep{bunne_unsb}, whose default configuration involves three time points, we selected samples at time points 0, 2, and 4 from the gene regulatory network dataset as inputs for the algorithms.}

\subsection{\revisenew{Hyperparameters Selection and Loss Weighting}}\label{appen:experdetails}
We use one A100 Nvidia GPU along with 16 CPU cores for computation at a shared high-performance computing cluster. \revise{For the two-stage training of DeepRUOT, each stage necessitates the selection of hyperparameters tailored to its specific requirements,} \revisenew{and here we will provide a guideline based on empirical studies}.

\revise{ In the pre-training phase, we need to select the parameters in $\mathcal{L}_{\text{Recons}}$, i.e., $\lambda_m$ and  $\lambda_d$ in $\mathcal{L}_{\text{Recons}}=\lambda_m\mathcal{L}_{\text{Mass}}+\lambda_d\mathcal{L}_{\text{OT}}$.
Parameter selection is important during this stage, as in the ablation studies we have shown that both components are important. \revisenew{To effectively manage these parameters, we implement a hyperparameter scheduling strategy during the pre-training stage. Initially, we assign predefined values to $\lambda_m$ and $\lambda_d$ and train the model for a specified number of epochs. Subsequently, we adjust $\lambda_m$ to zero ($\lambda_m = 0$) and conduct additional epochs of training. This strategy is grounded in the rationale of first aligning growth patterns and subsequently refining the transition process based on the established growth rates. }\revisenew{The parameter scheduling procedure is elucidated in  \cref{tab:parameter_settings}, where the arrow notation ($\rightarrow$) signifies the rescheduling of hyperparameter values during the pre-training stage. Specifically, each arrow indicates the change from an initial value to a subsequent value, reflecting the strategic adjustment of hyperparameters during the pre-training stage.}}

\revise{
During the \revisenew{training} stage, the pre-training phase provides robust initial values, thereby reducing the sensitivity to parameter variations in $
\mathcal{L}= \mathcal{L}_{\text{Energy}}+\lambda_r\mathcal{L}_{\text{Recons}}+\lambda_f\mathcal{L}_{\text{FP}}.
$ As a result, we utilize the set of parameters and train the model for  10  epochs consistently across different datasets.}

\revisenew{As shown in \cref{tab:parameter_settings}, the primary hyperparameters requiring adjustment during the pre-training stage are mainly the number of training epochs selected during the $\lambda_m$ resetting process, specifically the number of epochs trained before $\lambda_m$ resetting and the number of epochs trained after $\lambda_m$ resetting. As previously discussed, the pre-training stage is crucial for stabilizing the training process. Determining the number of epochs relies on the complexity and scale of the problem, including factors such as the number of time points and the size of the dataset. For the first epoch number, a default of 30 epochs is recommended. However, in the case of the Gene Regulatory Network dataset, we opted for 20 epochs due to its larger number of time points and greater data volume. Similarly, for the EMT dataset, 20 epochs were chosen as it had already achieved convergence.}
\begin{table}[htbp]
    \centering
    \caption{\revisenew{ Parameter Settings for Different Datasets Across Two Training Stages (Regulatory Network ($\sigma=0.25$), Mouse Hematopoiesis ($\sigma=0.25$), EMT ($\sigma=0.05$), Gaussian Mixtures ($\sigma=0.1$)). The arrow ($\rightarrow$) indicates the scheduling of parameter values within the Pre-Training Stage, where the numbers on the left indicate the hyperparameter values before resetting, and the right ones indicate the values after resetting. For example, the $(1.0, 0.1, 20)\rightarrow (0.0,0.1,10)$ entry in ($\lambda_m$, $\lambda_d$, Epochs) row denotes 20 epochs training when $\lambda_m$ was $1.0$, $\lambda_d$ was $0.1$, and another 10 epochs training after $\lambda_m$ was reset to $0.0$.}}
    \label{tab:parameter_settings}\resizebox{1\textwidth}{!}{
    \begin{tabular}{lccccccc}
        \toprule
        \multirow{2}{*}{\textbf{Parameter}} & \multicolumn{7}{c}{\textbf{Datasets}} \\
        \cmidrule(lr){2-8}
         & {\textbf{Regulatory Network}} & \textbf{Mouse Hematopoiesis} & \textbf{EMT } & \multicolumn{4}{c}{\textbf{Gaussian Mixtures}} \\
         &  &  &  & \textbf{10D} & \textbf{20D} & \textbf{50D} & \textbf{100D} \\
        \midrule
      
        \multicolumn{8}{l}{\textbf{Pre-Training Stage (With Hyperparameters  Scheduling)}} \\
        \midrule
        ($\lambda_m, \lambda_d$, Epochs)  & $(1.0, 0.1, 20) \rightarrow (0.0, 0.1, 10)$ & $(1.0, 0.1, 30) \rightarrow (0.0, 0.1, 30)$ & $(1.0, 0.1, 20) \rightarrow (0.0, 0.1, 0)$ & $(1.0, 0.1, 35) \rightarrow (0.0, 0.1, 80)$ & $(1.0, 0.1, 30) \rightarrow (0.0, 0.1, 90)$ & $(1.0, 0.1, 35) \rightarrow (0.0, 0.1, 120)$ & $(1.0, 0.1, 35) \rightarrow (0.0, 0.1, 100)$ \\
        \midrule
        \multicolumn{8}{l}{\revisenew{\textbf{Training Stage}}} \\
        \midrule
             $\lambda_m$ & $1 \times 10^{3}$ & $1 \times 10^{3}$ & $1 \times 10^{3}$ & $1 \times 10^{3}$ & $1 \times 10^{3}$ & $1 \times 10^{3}$ & $1 \times 10^{3}$ \\
        $\lambda_d$ & 1.0 & 1.0 & 1.0 & 1.0 & 1.0 & 1.0 & 1.0 \\
        $\lambda_r$ & 1.0 & 1.0 & 1.0 & 1.0 & 1.0 & 1.0 & 1.0 \\
        $\lambda_f$ & 1.0 & 1.0 & 1.0 & 1.0 & 1.0 & 1.0 & 1.0 \\
        $\lambda_w$ & \revisenew{0.1} & \revisenew{0.1} & \revisenew{0.1} & \revisenew{0.1} & \revisenew{0.1} & \revisenew{0.1} & \revisenew{0.1} \\
        Epochs & 10 & 10 & 10 & 10 & 10 & 10 & 10 \\
        \bottomrule
    \end{tabular}
    }
\end{table}

\revisenew{
In \cref{tab:epoch_scheduling}, we evaluated the GRN dataset by maintaining the second epoch number (trained after $\lambda_m$ resetting) as 10 while varying the first epoch number (trained before resetting) among 10, 20, and 30, with the other hyperparameters chosen as \cref{tab:parameter_settings}. The results indicated that the first epoch number value of 10, which was insufficient,  deteriorated performance, whereas the first epoch number values of 20 and 30 showed negligible differences. This suggests that increasing the number of pre-training epochs does not adversely affect the training process; hence, we recommend training for a higher number of epochs when feasible. The selection of the second epoch number varies across datasets based on convergence criteria. For instance, in the GRN dataset, training with the second epoch number set to 10 was sufficient for convergence, leading us to terminate further training. Conversely, for the EMT dataset, convergence was achieved during the first stage, negating the need for additional second epoch number training. Generally, a higher number of epochs in this stage enhances training efficacy. In the case of Gaussian Mixtures, the number of the second epoch number increases with dimensionality. However, for 50D and 100D, the second epoch number decreases because, for 100D, training can extend beyond 120 epochs for improved results, while 50D converges around 100 epochs, making further training unnecessary. Therefore, for the adjustment of the second epoch number, we recommend that users train until convergence based on their specific data requirements.
}

\begin{table}[htbp]
    \centering
    
    \caption{Results of different epoch scheduling during the pre-training stage on gene regulatory dataset. Here we use the arrow to represent the number of epochs trained before(left) and after(right) the $\lambda_m$ resetting, for example, $10\rightarrow 20$ denoting 10 epochs of training when $\lambda_m=1$ and 20 epochs of training when $\lambda_m=0$. }
    \begin{threeparttable}
    \resizebox{\textwidth}{!}{ % Resize table to fit the width of the text
    \begin{tabular}{lcccccccc}
    \toprule
    & \multicolumn{2}{c}{\boldmath $t=1$} & \multicolumn{2}{c}{\boldmath $t=2$} & \multicolumn{2}{c}{\boldmath $t=3$} & \multicolumn{2}{c}{\boldmath $t=4$} \\
    \cmidrule(lr){2-3} \cmidrule(lr){4-5} \cmidrule(lr){6-7} \cmidrule(lr){8-9}
    \textbf{Epoch Scheduling} & $\mathcal{W}_1$ & $\mathcal{W}_2$ & $\mathcal{W}_1$ & $\mathcal{W}_2$ & $\mathcal{W}_1$ & $\mathcal{W}_2$ & $\mathcal{W}_1$ & $\mathcal{W}_2$ \\
    \midrule
    10 $\rightarrow$ 10 & $0.138_{\pm 0.003}$ & $0.173_{\pm 0.002}$ & $0.243_{\pm 0.005}$ & $0.278_{\pm 0.004}$ & $0.325_{\pm 0.011}$ & $0.353_{\pm 0.013}$ & $0.406_{\pm 0.014}$ & $0.455_{\pm 0.011}$ \\
    20 $\rightarrow$ 10 & ${0.095}_{\pm 0.005}$ & ${0.115}_{\pm 0.006}$ & ${0.104}_{\pm 0.002}$ & ${0.130}_{\pm 0.007}$ & ${0.130}_{\pm 0.003}$ & ${0.157}_{\pm 0.003}$ & ${0.140}_{\pm 0.007}$ & ${0.168}_{\pm 0.007}$ \\
    30 $\rightarrow$ 10 & $0.107_{\pm 0.008}$ & $0.128_{\pm 0.010}$ & $0.120_{\pm 0.05}$ & $0.152_{\pm 0.08}$ & $0.130_{\pm 0.02}$ & $0.164_{\pm 0.004}$ & $0.154_{\pm 0.012}$ & $0.188_{\pm 0.113}$ \\
    \bottomrule
    \end{tabular}
    } % End of resizebox
    \end{threeparttable}
    \label{tab:epoch_scheduling}
\end{table}

\revisenew{In the training stage, we empirically identified $\lambda_w$ as a hyperparameter requiring tuning. We found that $\lambda_w$ may relate to the noise amplitude  $\sigma$. Specifically, for the Gene Regulatory Network (GRN) and Mouse Hematopoiesis datasets, where $\sigma$ is set to 0.25, a smaller $\lambda_w$ is necessary to maintain computational stability; setting $\lambda_w$ to 1 in these cases resulted in instability. For the remaining datasets, we have found that selecting $\lambda_w$ within the range of [0.1, 1] is feasible, as variations within this range do not significantly impact the training outcomes.
In \cref{table:emt_lambda_w_ablation}, we present the results of training with $\lambda_w$ values of 0.1, 0.4, 0.7, and 1.0 on EMT dataset. We computed the values of the trained results ($\vxt$, $g$, $s$) at each data point, using the results obtained with $\lambda_w = 1$ as the baseline. The distances between the outcomes of other $\lambda_w$ settings and the baseline were measured using Euclidean distance for inferred $\vxt$ and mean squared error (MSE) for $g$ and $s$. As illustrated in  \cref{table:emt_lambda_w_ablation}, we observed that as long as $\lambda_w$ ensures computational stability, the outcomes for different $\lambda_w$ values are consistent. Therefore, we recommend selecting a smaller $\lambda_w$, with a default value of 0.01 (for smaller $\sigma$) or 0.001 (for larger $\sigma$), to achieve stable and consistent training results.}

\begin{table}[htbp]
    \centering
    
    \caption{Results with different $\lambda_w$ settings on the EMT dataset ($\sigma=0.25$). Distances for $\boldsymbol{v}$ are measured using Euclidean distance, while $g$ and $s$ are evaluated using mean squared error (MSE).}
    \begin{threeparttable}
% Resize table to fit the width of the text
    \begin{tabular}{l|ccc}
    \toprule
    \textbf{$\lambda_w$} & $\boldsymbol{v}$ Dist & $g$ MSE & $s$ MSE \\
    \midrule
    0.1 & $2.64 \times 10^{-8}$ & $6.61 \times 10^{-16}$ & $1.63 \times 10^{-9} $ \\
    0.4 & $2.20 \times 10^{-8}$ & $3.80 \times 10^{-16}$ & $7.23 \times 10^{-10}$ \\
    0.7 & $2.11 \times 10^{-8}$ & $3.46 \times 10^{-16}$ & $1.80 \times 10^{-10} $ \\
    1.0 & $0.00$ & $0.00$ & $0.00$ \\
    \bottomrule
    \end{tabular}

    \end{threeparttable}
    \label{table:emt_lambda_w_ablation}
\end{table}

\subsection{Training Initial Log Density Function}\label{appen:initiallog}
 In our gene regulatory network example training, we utilize the following steps to stabilize training. 
Here since we parameterize $\frac{\sigma^2}{2}\log \pxt$ using $s_\theta (\boldsymbol{x}, t)$ when we compute Fokker-Planck constrained loss, we need to do the exponential operations, which may cause numerical instability during computations. To mitigate this issue, we augment the CFM loss with an additional penalty term defined as $L = \alpha_{\text{penalty}} \max(s_\theta (\boldsymbol{x}, t), 0)$ prior to training $s_\theta (\boldsymbol{x}, t)$. This penalty encourages $s_\theta (\boldsymbol{x}, t)$ to adopt negative values, thereby preventing instability caused by the subsequent exponential function. Our training procedure is conducted in two stages:

\begin{enumerate}
    \item \textbf{Initial Training Phase:} We optimize the combined CFM loss and penalty term for 3,000 epochs. This stage ensures that $s$ remains negative, stabilizing the training process by avoiding potential numerical issues associated with positive values of $s$.
    
    \item \textbf{Secondary Training Phase:} After the initial phase, we remove the penalty term and continue training solely with the CFM loss for an additional 6,000 epochs. This allows the model to fine-tune the parameter $s$ without the constraint imposed by the penalty, facilitating more accurate learning.
\end{enumerate}

For the last two examples, such stabilization measures are unnecessary, as the dynamics of these cases do not induce the same level of numerical instability. Consequently, we proceed with training using only the CFM loss without incorporating the penalty term.

This issue may be avoided if we use the Hamilton–Jacobi–Bellman (HJB) equations to set the constraints rather than the direct Fokker-Planck equations \citep{epr,LIWUCHEN_score,jiang2024physics,meng2024hj}.

\section{Technical Details for Schrödinger Bridge and RUOT}
\subsection{Proof of Dynamical Formulation of Schrödinger Bridge}\label{appen:proofdynaSB}
\begin{proof}
    For Schrödinger bridge problems, the optimal solution can be found within the class of SDEs:
$$
\Xt \sim \mu^\Xb_t: \quad \mathrm{d} \Xt=\boldsymbol{b}\left(\Xt, t\right) \mathrm{d} t+\boldsymbol{\sigma}\left(\Xt, t\right)\mathrm{d}\Wt ,
$$
The Fokker-Planck equation for the SDE  is
$$
\partial_t \pxt=-\nabla_{\boldsymbol{x}} \cdot\left(\pxt \bxt\right)+\frac{1}{2} \nabla_{\boldsymbol{x}}^2:\left(\axt \pxt\right),
$$ 
where $\axt=\dxt \dxT$.
Then we need to compute the KL divergence of these two stochastic processes. Define the auxiliary variable
$$\gamma(\xb,t)=\boldsymbol{\sigma}(\xb,t)^{-1}\boldsymbol{b}(\xb,t)$$
and apply the Girsanov’s theorem, we get
\begin{equation}
\label{eq:dsbkl}
\begin{aligned}
\mathcal{D}_{\text{KL}}\left(\mu^\Xb_{[0,1]}||\mu^\Yb_{[0,1]}\right) =\mathbb{E}_{\mathbb{P}} \left[ \mathcal{E}_{1}  \left(\int_{0}^{1} \gamma(\Yt,t)\rmd \Wt- \frac{1}{2}\int_0^1\left\|\gamma({\boldsymbol{Y}_t}, t)\right\|_2^2\rmd t\right)\right],
\end{aligned}
\end{equation}
where $\{\mathcal{E}_t\}_{t\in[0,1]}$ is an exponential martingale with respect to the Brownian filtrations $\{\mathcal{F}_t\}$ defined as
$$
\mathcal{E}_t=\exp  \left(\int_{0}^{t} \gamma(\Yt,t)\rmd \Wt- \frac{1}{2}\int_0^t\left\|\gamma({\boldsymbol{Y}_t}, t)\right\|_2^2\rmd t\right).
$$
It is well-known that the exponential martingale $\mathcal{E}_t$ satisfies the SDE
$$
\rmd \mathcal{E}_t=\mathcal{E}_t \gamma(\Yt,t) \rmd\Wt,\quad \mathcal{E}_0=1.
$$
So we obtain
$$
\mathcal{E}_1=\int_{0}^{1} \mathcal{E}_s \gamma(\Ys,s)\rmd \Ws+1.
$$
Since $\mathbb{E}\int_{0}^{1}\gamma(\Yt,t)\rmd \Wt=0$, we get
$$
\begin{aligned}
\mathbb{E}\left[\mathcal{E}_{1} \int_{0}^{1}\gamma(\Yt,t)\rmd \Wt\right]&=\mathbb{E} \left[\int_{0}^{1} \mathcal{E}_t \gamma(\boldsymbol{Y}_t,t) \rmd \Wt\cdot \int_{0}^{1}\gamma(\boldsymbol{Y}_t,t) \rmd \Wt\right],\\
&=\mathbb{E}\int_{0}^{1}\mathcal{E}_t\left\|\gamma(\boldsymbol{Y}_t, t)\right\|_2^2 \rmd t,\\
&=\mathbb{E}\int_{0}^{1} \mathbb{E}\left(\left\|\gamma(\boldsymbol{Y}_t, t)\right\|_2^2\mathcal{E}_1\mid \mathcal{F}_t\right) \rmd t,\\
&=\int_{0}^{1}\mathbb{E}\left(\mathcal{E}_1\left\|\gamma(\boldsymbol{Y}_t, t)\right\|_2^2\right) \rmd t, \\
&= \mathbb{E}\left[\mathcal{E}_1\int_{0}^{1}\left\|\gamma(\boldsymbol{Y}_t, t)\right\|_2^2 \rmd t\right].
\end{aligned}
$$
Combine with \eqqref{eq:dsbkl}, we obtain that 
$$
\begin{aligned}
\mathcal{D}_{\text{KL}}(\mu^\Xb_{[0,1]}||\mu^\Yb_{[0,1]})&=\mathbb{E}\left[ \mathcal{E}_{1} \left(\int_{0}^{1}\frac{1}{2}\left\|\gamma(\boldsymbol{Y}_t, t)\right\|_2^2\rmd t\right)\right],\\
&=\mathbb{E}\int_{0}^{1}\frac{1}{2}\left\|\gamma(\boldsymbol{X}_t, t)\right\|_2^2 \rmd t,\\
&=\int_{0}^{1}\int_{\mathbb{R}^d}\left[\frac{1}{2}\boldsymbol{b^T}(\boldsymbol{x}, t)\axt^{-1}\boldsymbol{b}(\boldsymbol{x}, t)\right]\pxt \rmd \boldsymbol{x}\rmd t.
\end{aligned}
$$
The proof is done.
\end{proof}
\subsection{Proof of Fisher Regularization of RUOT}\label{appen:fisherRUOT}
\begin{proof}
From \eqref{eq:undsb3} we obtain
$$
\begin{aligned}
\partial_t p&=-\nabla_{\boldsymbol{x}} \cdot\left(p \boldsymbol{b}\right)+\frac{1}{2}\nabla_{\boldsymbol{x}}^2: \left(\sigma^2(t)\boldsymbol{I}p\right) + g(\boldsymbol{x},t)p, \\
&=- \nabla_{\boldsymbol{x}} \cdot\left(\left(\boldsymbol{b}- \frac{1}{2} \sigma^2(t)\nabla_{\boldsymbol{x}} \log p \right)p\right)+ \gxt p. \\
\end{aligned}
$$
Using the change of variable $\vxt=\bxt-\frac{1}{2} \sigma^2(t)\nabla_{\boldsymbol{x}} \log p $, we see that it is equivalent to
$$
\partial_t p=-\nabla_{\boldsymbol{x}} \cdot\left(p \vxt\right) + g(\boldsymbol{x},t) p .
$$
Correspondingly, the integrand in the objective functional becomes
\begin{equation}\label{eqn:laventfisher}
\int_0^1\int_{\Rd}\left[ \frac{1}{2}\left\|\vxt\right\|_2^2+\frac{\sigma^4(t)}{8}\left\|\nabla_{\boldsymbol{x}}\log p\right\|_2^2+\frac{1}{2}\left\langle \vxt, \sigma^2(t)\nabla_{\boldsymbol{x}} \log p\right\rangle + \alpha \Psi\left(g\right)\right]\pxt \rmdx \mathrm{d} t.
\end{equation}

Letting $H\left(\pxt\right):=\int_{\mathbb{R}^d} \pxt \log \pxt \rmd \boldsymbol{x}$ be the entropy, we have
$$
\begin{aligned}
&\sigma^2(1)H\left(p(\boldsymbol{x}, 1)\right)-\sigma^2(0)H\left(p(\boldsymbol{x}, 0\right)  =\int_0^1 \partial_t \left(\sigma^2(t)H\left(\pxt\right)\right) \mathrm{d} t, \\
& =\int_0^1 \int_{\Rd}\frac{\rmd \sigma^2(t)}{\rmd t}\pxt \log \pxt + \sigma^2(t) \left(1+\log \pxt\right) \partial_t \pxt \rmdx \mathrm{d} t, \\
& =\int_0^1 \int_{\Rd}\sigma^2(t)\left(1+\log \pxt\right) \cdot\left(-\nabla_{\boldsymbol{x}} \cdot\left(\pxt\vxt\right)+ g \pxt\right)+\frac{\rmd \sigma^2(t)}{\rmd t}\pxt \log \pxt  \rmdx \mathrm{d} t, \\
& =\int_0^1 \int_{\Rd}\sigma^2(t)\pxt\left\langle\nabla_{\boldsymbol{x}} \log \pxt, \vxt\right\rangle + \sigma^2(t)\left(1+\log \pxt\right)gp +\frac{\rmd \sigma^2(t)}{\rmd t}\pxt \log \pxt   \rmdx \mathrm{d} t .
\end{aligned}
$$
Therefore,
$$
\begin{aligned}
    \int_0^1\int_{\Rd}\left[\left\langle \sigma^2(t) \nabla_{\boldsymbol{x}} \log p, \vxt\right\rangle\right]\pxt \rmdx \mathrm{d} t&=\left(\sigma^2(1)H\left(p(\boldsymbol{x}, 1)\right)-\sigma^2(0)H\left(p(\boldsymbol{x}, 0)\right)\right)\\&-\int_0^1\int_{\Rd}\sigma^2(t)\left(1+\log \pxt\right)g  p\rmdx \mathrm{d} t \\
    &-\int_0^1\int_{\Rd}\frac{\rmd \sigma^2(t)}{\rmd t}\pxt \log \pxt\rmdx \mathrm{d} t.
    \end{aligned}
$$
 \end{proof}
 \subsection{RUOT with Nonlinear Fokker-Planck Equation Constraints}\label{appen:RUOTnonlinear}
Note that when $\Psi\left(\gxt\right)$ take the quadratic form i.e. $\Psi\left(\gxt\right)=|\gxt|_2^2$, then by canceling the cross term in \cref{thm:unddsb_fisher} we will get another Fisher information regularisation form as below.
\begin{theorem}
\label{thm:unddsb_fisher_wfr}
When $\Psi\left(\gxt\right)=|\gxt|_2^2$ and $\sigma(t)=\sigma$ is constant, the  regularized unbalanced optimal transport problem \eqref{eq:unddsb_fisher} is equivalent to 
\begin{equation}
     \inf_{\left(p, \boldsymbol{v}, \widetilde{g}\right)} \int_0^1\int_{\Rd}\left[ \frac{1}{2}\left\|\vxt\right\|_2^2+\frac{\sigma^4}{8}\left\|\nabla_{\boldsymbol{x}} \log p\right\|_2^2+\alpha |\tgxt|^2-\frac{\sigma^4}{16\alpha}(1+\log p)^2 \right]p\rmdx\mathrm{d} t,
\end{equation}
where the infimum is taken all pairs $\left(p, \boldsymbol{v}, \widetilde{g}\right)$ such that $p(\cdot , 0)=$ $\nu_0, p(\cdot , 1)=\nu_1, p(\boldsymbol{x},t)$ absolutely continuous, and
\begin{equation}
    \label{eq:unddsb_fisher_wfr}
 \partial_t p=-\nabla_{\boldsymbol{x}} \cdot\left(p \vxt\right) + \tgxt p+ \frac{\sigma^2}{4 \alpha}(1+\log p)p .
\end{equation}  
 coupled with vanishing boundary condition: $\displaystyle \lim_{|\boldsymbol{x}| \rightarrow \infty}p(\boldsymbol{x},t)=0$.
 \end{theorem}
 \begin{proof}
     Based on the formulation of \cref{thm:unddsb_fisher}, we proceed by introducing the following change of variables:
     \begin{equation}
         \widetilde{g}=g-\frac{\sigma^2}{4 \alpha}(1+\log p).
     \end{equation}
     Substituting this variable into \eqqref{eq:unddsb_fisher}, then it follows immediately.
 \end{proof}
 \vspace{0.5em}

   From \cref{thm:unddsb_fisher_wfr}, we can find that if we introduce a new term in the continuity equation \eqqref{eq:undsb3} as:
 $$
 \partial_t p=-\nabla_{\boldsymbol{x}} \cdot\left(p \boldsymbol{b}\right)+\frac{1}{2}\nabla_{\boldsymbol{x}}^2: \left(\sigma^2\boldsymbol{I}p\right)+gp-\frac{\sigma^2}{4 \alpha}(1+\log p)p.
 $$
Then \eqqref{eq:unddsb_fisher_wfr} now yields the standard continuity equation:
$$
 \partial_t p=-\nabla_{\boldsymbol{x}} \cdot\left(p \vxt\right) + \tgxt p.
$$  
So we introduce a new definition. 
\begin{definition}
\label{def:duot}
    We introduce
$$
\operatorname{UOT}^{\mathrm{D}}\left(\nu_0, \nu_1\right):=\inf _{(p, \boldsymbol{b}, g) \in \mathcal{C}\left(\nu_0, \nu_1\right)}\int_0^1 \int_{\Rd}\frac{1}{2}\left\|\bxt\right\|_2^2p\rmdx\rmd t+\int_0^1\int_{\Rd} \alpha \left|g(\boldsymbol{x},t)\right|_2^2 p \rmdx\mathrm{d} t,
$$
where, for $(p, \boldsymbol{b}, g)$ belonging to appropriate functions spaces,
$$
\mathcal{C}\left(\nu_0, \nu_1\right):=\left\{p, \boldsymbol{b}, g\left| \partial_t p=-\nabla_{\boldsymbol{x}} \cdot\left(p \boldsymbol{b}\right)+\frac{1}{2}\nabla_{\boldsymbol{x}}^2: \left(\sigma^2\boldsymbol{I}p\right)+gp-\frac{\sigma^2}{4 \alpha}(1+\log p)p\right|\right\},
$$
where $p_0=\nu_0$ and $p_1=\nu_1$.
\end{definition}
  \vspace{0.5em}
 \begin{remark}
    Note that \cref{def:duot}  is consistent with the form provided in \citep{highorderRUOT}, and it has been conjectured that it is the higher order approximation scheme to the unbalanced optimal transport problem as $\sigma \rightarrow 0$. 
 \end{remark}
  \vspace{0.5em}
  
\subsection{RUOT in the General Case}\label{appen:RUOTgeneral}
For the general case, we consider the following equation
    \begin{equation}
        \label{eq:gen_undsb1}
\partial_t p=-\nabla_{\boldsymbol{x}} \cdot\left(p \boldsymbol{b}\right)+\frac{1}{2}\nabla_{\boldsymbol{x}}^2: \left(\axt p\right)+gp,
    \end{equation}
      with the corresponding metric 
    \begin{equation}
        \label{eq:gen_undsb2}
\int_0^1\int_{\Rd} \frac{1}{2}\left\|\bxt\right\|_2^2p(\boldsymbol{x},t)\rmdx\mathrm{d} t+ \int_0^1\int_{\Rd} \alpha \Psi\left(g(\boldsymbol{x},t)\right) p(\boldsymbol{x},t)   \rmdx\mathrm{d} t,
    \end{equation}
    where $\alpha$ and $\Psi: \mathbb{R} \rightarrow [0, +\infty]$ is the growth penalization.
   Then we can define the 
    \emph{regularized unbalanced optimal transport} problem in the same way.
     \begin{definition}[Regularized unbalanced optimal transport II]
    \label{def:gen_unddsb}
     Consider
       \begin{equation}
\inf _{\left(p, \boldsymbol{b}, g\right)} \int_0^1\int_{\Rd} \frac{1}{2}\left\|\bxt\right\|_2^2\pxt\rmdx\mathrm{d} t+ \int_0^1\int_{\Rd} \alpha \Psi\left(g(\boldsymbol{x},t)\right) \pxt  \rmdx\mathrm{d} t,
    \end{equation}
    where the infimum is taken all pairs $\left(p, \boldsymbol{b}\right)$ such that $p(\cdot , 0)=$ $\nu_0, p(\cdot , 1)=\nu_1, p(\boldsymbol{x},t)$ absolutely continuous, and
    \begin{equation}
        \label{eq:gen_undsb3}
\partial_t p=-\nabla_{\boldsymbol{x}} \cdot\left(p \boldsymbol{b}\right)+\frac{1}{2}\nabla_{\boldsymbol{x}}^2: \left(\axt p\right)+gp  .
    \end{equation}
    coupled with vanishing boundary condition: $\displaystyle \lim_{|\boldsymbol{x}| \rightarrow \infty}p(\boldsymbol{x},t)=0$.
    \end{definition}
    
    We similarly can
reformulate \cref{def:gen_unddsb} with the following Fisher information regularization.
\begin{theorem}
\label{thm:gen_unddsb_fisher}
    Consider the  regularized unbalanced optimal transport problem \cref{def:gen_unddsb}, it is equivalent to 
\begin{equation}
\label{eq:gen_unddsb_fisher}
\begin{aligned}
    &\inf _{\left(p, \boldsymbol{v},g\right)} \int_0^1\int_{\Rd} \frac{1}{2}\left\|\vxt\right\|_2^2\pxt\rmdx\mathrm{d} t+\frac{1}{8}\left\|\dxt\nabla_{\boldsymbol{x}}\cdot \log \left(\axt p\right)\right\|_2^2\pxt\rmdx\mathrm{d} t\\&+\frac{1}{2}\left\langle \vxt, \axt\nabla_{\boldsymbol{x}}\cdot \log \left(\axt p\right)\right\rangle\pxt\rmdx\mathrm{d} t+ \alpha \Psi\left(g(\boldsymbol{x},t)\right) \pxt\rmdx\mathrm{d} t,
\end{aligned}
\end{equation}
where the infimum is taken all pairs $\left(p, \boldsymbol{v}, g\right)$ such that $p(\cdot , 0)=$ $\nu_0, p(\cdot , 1)=\nu_1, p(\boldsymbol{x},t)$ absolutely continuous, and
\begin{equation}
\label{eq:gen_unddsb_fisher_contineq}
    \partial_t p=-\nabla_{\boldsymbol{x}} \cdot\left(p \vxt\right) + g(\boldsymbol{x},t) p .
\end{equation}
 coupled with vanishing boundary condition: $\displaystyle \lim_{|\boldsymbol{x}| \rightarrow \infty}p(\boldsymbol{x},t)=0$.
 \end{theorem}
 \begin{proof}
From \cref{def:gen_unddsb} we know that
the Fokker-Planck equation for the SDE  is
$$
\begin{aligned}
\partial_t p&=-\nabla_{\boldsymbol{x}} \cdot\left(p \boldsymbol{b}\right)+\frac{1}{2}\nabla_{\boldsymbol{x}}^2: \left(\axt p\right) + g(\boldsymbol{x},t)p, \\
&=- \nabla_{\boldsymbol{x}} \cdot\left(\left(\boldsymbol{b}- \frac{1}{2} \axt\nabla_{\boldsymbol{x}}\cdot \log \left(\axt p\right) \right)p\right)+ \gxt p. \\
\end{aligned}
$$
with minimization problem 
$$
\inf _{\left(p, \boldsymbol{b}\right)} \int_0^1\int_{\Rd} \frac{1}{2}\left\|\bxt\right\|_2^2\pxt\rmdx\mathrm{d} t+ \int_0^1\int_{\Rd} \alpha \Psi\left(g(\boldsymbol{x},t)\right)    \pxt\rmdx\mathrm{d} t.
$$
Using the change of variable $\vxt=\bxt-\frac{1}{2} \axt\nabla_{\boldsymbol{x}}\cdot \log \left(\axt p\right) $, we see that it is equivalent to
$$
\partial_t p=-\nabla_{\boldsymbol{x}} \cdot\left(p \vxt\right) + g(\boldsymbol{x},t) p .
$$

On the other hand, since $\left\|\bxt\right\|_2^2=\left\|\vxt\right\|_2^2+\frac{1}{4}\left\|\axt\nabla_{\boldsymbol{x}}\cdot \log \left(\axt p\right)\right\|_2^2+2\left\langle \vxt, \frac{1}{2}\axt\nabla_{\boldsymbol{x}}\cdot \log \left(\axt p\right)\right\rangle$, the integrand in the objective then becomes
$$
\begin{aligned}
    &\int_0^1\int_{\Rd} \frac{1}{2}\left\|\vxt\right\|_2^2\pxt\rmdx\mathrm{d} t+\frac{1}{8}\left\|\dxt\nabla_{\boldsymbol{x}}\cdot \log \left(\axt p\right)\right\|_2^2\pxt\rmdx\mathrm{d} t\\&+\frac{1}{2}\left\langle \vxt, \axt\nabla_{\boldsymbol{x}}\cdot \log \left(\axt p\right)\right\rangle\pxt\rmdx\mathrm{d} t+ \alpha \Psi\left(g(\boldsymbol{x},t)\right) \pxt\rmdx\mathrm{d} t,
\end{aligned}
$$
 \end{proof}

  Similarly, from the proof of \cref{thm:gen_unddsb_fisher},  the original SDE
  $\mathrm{d} \Xt=\left(\boldsymbol{b}\left(\Xt, t\right)\right) \mathrm{d} t+\boldsymbol{\sigma}\left( \boldsymbol{X}_t, t\right)\mathrm{d}\Wt$
  can be transformed 
  into the probability flow ODE
  $$
  \mathrm{d} \Xt=\underbrace{\left(\boldsymbol{b}\left(\Xt, t\right)-\frac{1}{2} \aXt\nabla_{\boldsymbol{x}}\cdot \log \left(\aXt \boldsymbol{p}\left(\Xt, t\right)\right)\right)}_{\boldsymbol{v}\left(\Xt, t\right)} \rmd t.
  $$
  Conversely, if the probability flow ODE’s drift $\vxt$ the diffusion rate $\axt$ and the function $\nabla_{\boldsymbol{x}}\cdot \log \left(\axt \pxt\right)$ are known, then the
the SDE’s drift term $b(\boldsymbol{x},t)$ can be determined by
$$
b(\boldsymbol{x},t)=\vxt+\frac{1}{2}\axt \nabla_{\boldsymbol{x}} \cdot \log \left(\axt \pxt\right).
$$
So to  specify an SDE is equivalent to specifying the probability flow ODE and the corresponding log density function.

\section{Connections with Schrödinger Bridge}\label{appen:connectionSB}
Although the RUOT problem can be derived from the dynamical formulation of a Schrödinger bridge relaxation, the relationship between the  RUOT problem and the original static Schrödinger bridge problem, the choice of the growth penalty form in RUOT as well as its microscopic interpretation, remains unsolved. In this section, we will discuss two potential microscopic interpretations with the choice of the growth penalty form and identify the persisting problems and challenges associated with each.
\subsection{Connections with Branching Schrödinger Bridge}
In \citep{branRUOT}, the authors explore the relationship between the RUOT and the branching Schrödinger bridge, with the corresponding reference process identified as branching Brownian motion. We will begin by reviewing their principal findings and subsequently highlight the associated issues and challenges.

A Branching Brownian Motion (BBM) can be formally described by the diffusivity  $\varepsilon$, branching rate  $\lambda$ (Each particle is associated with an independent exponential distribution with parameter  $\lambda$), and the offspring distribution denoted by  $\boldsymbol{p} = (p_k)_{k \in \mathbb{N}} \in \mathcal{P}(\mathbb{N})$ (The probability measures on  $\mathbb{N}$), where  $p_k$  is the probability of producing  $k$  offspring at a branching event. We define $ q_k = \lambda p_k $ and introduce the generating function of  $\boldsymbol{q}$  as  $\psi_{\boldsymbol{q}}$.
$$
\psi_{\boldsymbol{q}}({z}):=\sum_{k \in \mathbb{N}} q_k z^k=\lambda \sum_{k \in \mathbb{N}} p_k z^k.
$$

The branching mechanism is characterized by  $\boldsymbol{q}$, since $\sum_{k \in \mathbb{N}} p_k =1$ and  $\lambda$,  $\boldsymbol{p}$  can be recovered from  $\boldsymbol{q}$  by  $\lambda =  \sum_{k \in \mathbb{N}} q_k$ and  $p_k = \lambda^{-1} q_k$  for  $k \in \mathbb{N}$. Therefore, it is sufficient to fully determine the BBM with  $\nu$,  $\boldsymbol{q} = \lambda \boldsymbol{p}$, and initial distribution $\nu_0$. So $R \sim \text{BBM}(\varepsilon, \boldsymbol{q}, \nu_0)$ represent such BBM. Their core result indicates that consider \cref{def:unddsb} where $\sigma(t)=\sigma$ is constant and $\varepsilon=\sigma^2$ then if the growth penalty term is defined as follows, it can correspond to a branching Schrödinger process.
\begin{equation}
\label{eq:growth_psi}
    \Psi_{\varepsilon, \boldsymbol{q}}^*(s)=\varepsilon\left(\psi_{\boldsymbol{q}}\left(e^{s / \varepsilon}\right) e^{-s / \varepsilon}-\psi_{\boldsymbol{q}}(1)\right)=\varepsilon \sum_{k=0}^{+\infty} q_k\left\{\exp \left((k-1) \frac{s}{\varepsilon}\right)-1\right\},
\end{equation}
and we define $\Psi_{\varepsilon, \boldsymbol{q}}(g)=\sup _{s \in \mathbb{R}} g s-\Psi_{\varepsilon, \boldsymbol{q}}^*(s)$. So by following their result, we can present some examples.
\paragraph{The case of only loss mass} This example has been included in their work. We consider $\varepsilon=\lambda=1$, $p_0=1$, and $\sum_{k\geq2}p_k=0$, which means the particle can only die at the branching event. Substituting it into \eqqref{eq:growth_psi}, we have
$$
  \Psi_{1, \boldsymbol{q}}^*(s)=\exp \left(-s\right)-1.
$$
Then by computing its Legendre transform we obtain
$$
\Psi_{1, \boldsymbol{q}}(g)=1+g-g\log (-g),
$$
where $g<0$. This form is consistent with the formula discussed in \citep{chen2022most}.
\paragraph{The case of only gain mass} Similarly, if we consider $\varepsilon=\lambda=1$, $p_2=1$, which means the particle can only divide into two parts at the branching event. Then by a similar calculation, we obtain
$$
\Psi_{1, \boldsymbol{q}}(g)=1-g+g\log (g),
$$
which $g>0$.
\paragraph{The case of both gain mass and loss mass}  This is also presented in their work. We consider $\varepsilon=\lambda=1$, $p_0=p_2=\frac{1}{2}$, which means the particle can divide into two parts or die at the same probability at the branching event, which we have
$$
\Psi_{1, \boldsymbol{q}}^*(s)=\cosh(s)-1.
$$
However in this case the Legendre transform would be more complicated but it has some asymptotic properties discussed in their work.

\paragraph{General case} Although we have illustrated several examples, challenges persist, particularly in the context of more general scenarios. Specifically, when the growth penalty term cannot be expressed as \(\Psi_{\varepsilon, \boldsymbol{q}}\), the connection to the corresponding stochastic process remains unclear. For instance, this ambiguity arises in cases where in the RUOT the $\Psi(g)$ takes the quadratic form i.e., the WFR metric, or when $\Psi(g)$ is in a linear form, such as $|g|$. These issues still require specific mathematical treatment and to our knowledge, remain an open problem.

\subsection{Connections with Schrödinger Bridge on Weighted Path} From an alternative perspective on the microscopic interpretation of RUOT, we aim to consider particles on the weighted path space, namely:
$$
\rmd \mu^{\boldsymbol{G}}_{[0,1]}=\exp\left(\int_{0}^{1}g(\boldsymbol{X}_s, s)\rmd s\right)\rmd \mu^{\boldsymbol{X}}_{[0,1]},
$$
where $\mu^{\boldsymbol{X}}_{[0,1]}$ represents the probability measure induced by the SDE (\eqqref{eq:sde}).
Here for example we consider a simple case when $g(\boldsymbol{x},t)=C$. Then following the procedure of \cref{thm:thm_fisher_simple}.
\begin{equation}
\begin{aligned}
\mathcal{D}_{\text{KL}}(\mu^{\boldsymbol{G}}_{[0,1]}||\mu^\Yb_{[0,1]})&=\mathbb{E}_{\mu^{\boldsymbol{G}}_{[0,1]}}  \left(\int_{0}^{1} \gamma({\boldsymbol{Y}_t}, t)\rmd \Wt- \frac{1}{2}\left\|\gamma({\boldsymbol{Y}_t}, t)\right\|_2^2\rmd t + C\right), \\
&=\mathbb{E} \left(\mathcal{E}_{1} \exp(C)\right)\left(\int_{0}^{1}\frac{1}{2}\left\|\gamma({\boldsymbol{Y}_t}, t)\right\|_2^2\rmd t\right)+C\exp(C),\\
&=\mathbb{E}\exp(C)\int_{0}^{1}\frac{1}{2}\left\|\gamma({\boldsymbol{X}_t}, t)\right\|_2^2 \rmd t+C\exp(C).
\end{aligned}
\end{equation}
Here $\mu^\Yb_{[0,1]}$ is the distribution of  stochastic process induced by \(\rmd \boldsymbol{Y}_t = \boldsymbol{\sigma}(\boldsymbol{Y}_t, t) \rmd \Wt\).
Our derivation reveals that, from a path weighting perspective, even when the growth term is a constant, it still does not correspond to the form of RUOT. This suggests that the microscopic interpretation from this angle may need further investigation to determine whether a RUOT problem could be recovered from this starting point.

\end{document}